\documentclass[preprint,authoryear]{elsarticle}

\usepackage{ycviu} 
\usepackage{framed,multirow}

\usepackage{amssymb}
\usepackage{amsmath}
\usepackage{latexsym}

\usepackage[utf8]{inputenc}
\usepackage[T1]{fontenc}
\usepackage{url}
\usepackage[dvipsnames]{xcolor}
\usepackage{todonotes}
\usepackage{tabularx}
\usepackage{booktabs}
\usepackage{multirow}
\usepackage{arydshln}

\usepackage{flushend}

\usepackage[position=t]{subcaption}
\usepackage[pagebackref=true,breaklinks=true,letterpaper=true,colorlinks,bookmarks=false]{hyperref}
\usepackage[capitalize]{cleveref}

\newcolumntype{Y}{>{\centering\arraybackslash}X}

\journal{Computer Vision and Image Understanding}

\begin{document}

\thispagestyle{empty}
                                                             
\begin{frontmatter}

\title{Distance transform regression for spatially-aware deep semantic segmentation}

\author[1,2]{Nicolas \snm{Audebert}\corref{cor1}} 
\cortext[cor1]{Corresponding author:} 
\ead{nicolas.audebert@onera.fr}
\author[1]{Alexandre \snm{Boulch}}
\author[1]{Bertrand \snm{Le Saux}}
\author[2]{Sébastien \snm{Lefèvre}}

\address[1]{DTIS, ONERA, Université Paris Saclay, F-91123 Palaiseau - France}
\address[2]{Univ. Bretagne-Sud, UMR 6074, IRISA, F-56000 Vannes, France}


\begin{abstract}
Understanding visual scenes relies more and more on dense pixel-wise classification obtained via deep fully convolutional neural networks. However, due to the nature of the networks, predictions often suffer from blurry boundaries and ill-segmented shapes, fueling the need for post-processing. This work introduces a new semantic segmentation regularization based on the regression of a distance transform. After computing the distance transform on the label masks, we train a FCN in a multi-task setting in both discrete and continuous spaces by learning jointly classification and distance regression. This requires almost no modification of the network structure and adds a very low overhead to the training process. Learning to approximate the distance transform back-propagates spatial cues that implicitly regularizes the segmentation. We validate this technique with several architectures on various datasets, and we show significant improvements compared to competitive baselines.
\end{abstract}

\end{frontmatter}


\section{Introduction}

Semantic segmentation is a task that is of paramount importance for visual scene understanding. It is often used as the first layer to obtain representations of a scene with a high level of abstraction, such as listing objects and their shapes.
Fully Convolutional Networks (FCNs) have proved themselves to be very effective for semantic segmentation of all kinds of images, from multimedia images~\cite{everingham_pascal_2014} to remote sensing data~\cite{rottensteiner_isprs_2012}, medical imaging~\cite{ulman_objective_2017} and autonomous driving~\cite{cordts_cityscapes_2016}.
However, a recurrent issue often raised by the practitioners is the fact that FCN tend to produce blurry or noisy segmentations, in which spatial transitions between classes are not as sharp as expected and objects sometimes lack connectivity or convexity, and therefore the results need to be regularized using some post-processing~\cite{zheng_conditional_2015,chen_deeplab_2018}.
This has led the computer vision community to investigate many post-processing and regularization techniques to sharpen the visual boundaries and enforce spatial smoothness in semantic maps inferred by FCN. Yet these methods are often either graphical models added on top of deep neural networks~\cite{liu_deep_2018,zheng_conditional_2015} or based on sophisticated prior knowledge~\cite{le_reformulating_2018,bertasius_semantic_2016}.
In this work, we propose a much straightforward approach by introducing a simple implicit regularization embedded in the network loss function. We consider the distance transform of the segmentation masks in a regression problem as a proxy for the semantic segmentation task.
The distance transform is a continuous representation of the label masks, where one pixel becomes represented not only by its belonging to a class, but by its spatial proximity to all classes. This means that the gradient back-propagated contains more information about the underlying spatial structure of the data compared to traditional classification.
As such, the network learns a smoother segmentation with a very low complexity overhead.
Moreover, this is straightforward to implement and does not rely on any additional priors, but only on an alternative representation of the ground truth. Therefore any deep segmentation architecture can be adapted in this fashion without any structural alteration.
We validate our method with several architectures on diverse application domains on which we obtain significant improvements w.r.t strong baselines: urban scene understanding, RGB-D images and Earth Observation.

\section{Related work}

Semantic segmentation is a longstanding task in the computer vision community. Several benchmarks have been introduced on many application domains such as COCO~\cite{lin_microsoft_2014} and Pascal VOC~\cite{everingham_pascal_2014} for multimedia images, CamVid~\cite{brostow_semantic_2009} and Cityscapes~\cite{cordts_cityscapes_2016} for autonomous driving, the ISPRS Semantic Labeling~\cite{rottensteiner_isprs_2012} and INRIA Aerial Image Labeling~\cite{maggiori_can_2017} datasets for aerial image, and medical datasets~\cite{ulman_objective_2017}, which are now dominated by the deep fully convolutional networks.
Many applications rely on a pixel-wise semantic labeling to perform scene understanding, such as object-instance detection and segmentation~\cite{he_mask_2017,arnab_pixelwise_2017} in multimedia images, segment-before-detect pipelines for remote sensing data processing ~\cite{audebert_segment-before-detect_2017,sommer_semantic_2017} and segmentation of medical images for neural structure detection and gland segmentation~\cite{ronneberger_u-net_2015,chen_dcan_2016}.

State-of-the-art architectures are all derived from the Fully Convolutional Network paradigm~\cite{long_fully_2015}, which introduced the possibility to perform pixel-wise classification using convolutional networks that were previously restricted to image-wide classification.
Many models building upon this idea were then proposed, e.g. DeepLab~\cite{chen_deeplab_2018}, dilated convolutional networks~\cite{yu_multi-scale_2015} or auto-encoder inspired architectures such as SegNet~\cite{badrinarayanan_segnet_2017} and U-Net~\cite{ronneberger_u-net_2015}.
The introduction of the residual learning framework~\cite{he_deep_2016} also introduced many new models for semantic segmentation, most notably the PSPNet~\cite{zhao_pyramid_2017} that incorporates multi-scale context in the final classification using a pyramidal module. 

However, one common deficiency of the FCNs is the lack of spatial-awareness that adversely affects the classification maps and makes spatial regularization a still active field of research~\cite{garcia-garcia_review_2017}. Indeed, predictions often tend to be blurry along the object edges. As FCN perform pixel-wise classification where all pixels are independently classified, spatial structure is fundamentally implicit and relies only on the use of convolutional filters. Although this has given excellent results on many datasets, this often leads to noisy segmentations, where artifacts might arise in the form of a lack of connectivity of objects and even salt-and-pepper noise in the classifications. Those problems are especially critical in remote sensing applications, in which most objects are fundamentally groups of convex
structures and where connectivity and inter-class transitions are a requirement for better mapping.

To address this issue, several approaches for smoothing have been suggested. Graphical models methods, such as dense Conditional Random Fields (CRF), have been used to spatially regularize the segmentation and sharpen the boundaries as a post-processing step~\cite{lin_efficient_2016}. However, this broke the end-to-end learning paradigm, and led to several reformulations in order to couple more tightly the graphical models with deep  networks. To this end, \cite{zheng_conditional_2015,liu_deep_2018} rewrote respectively the Conditional Random Field (CRF) and the Markov Random Field (MRF) graphical models as trainable neural networks. In a similar concept, \cite{le_reformulating_2018} reformulates the Variational Level Set method to solve it using an FCN, while~\cite{chen_semantic_2016} uses CNN to perform domain transform filtering. Those methods all are revisiting traditional vision techniques adapted to fit into the deep learning framework. However, they require heavy network modification and are computationally expensive.

A more straightforward strategy consists in performing a data-driven regularization by enforcing new constraints on the model in the form a special loss penalty. Notably, this area has been investigated in the literature for edge detection~\cite{yang_object_2016}. For instance, \cite{kokkinos_pushing_2015,bertasius_semantic_2016} introduce a carefully crafted loss especially tailored for object boundary detection. CASENet~\cite{yu_casenet_2017} tries to leverage semantics-related priors into the edge prediction task by learning the classes that are adjacent to each boundary, while the COB strategy~\cite{maninis_convolutional_2018} incorporates geometric cues by predicting oriented boundaries.
Multi-scale approaches such as~\cite{liu_richer_2017} tune the network architecture to fuse activations from multiple layers and improve edge prediction by mixing low and high-level features.
In the case of semantic segmentation, object shapes benefit from a better spatial regularity as most shapes are often clean, closed sets. Therefore, better boundaries often help by closing the contours and removing classification noise. To this end, models such as DeepContours~\cite{shen_deepcontour_2015} explicitly learn both the segmentation and the region boundaries using a multi-task hand-crafted loss. A similar approach with an ensemble of models has been suggested in~\cite{marmanis_classification_2017}, especially tailored for aerial images. ~\cite{chen_dcan_2016,cheng_fusionnet_2017} still use a multi-task loss with explicit edge detection, but also fuse feature maps from several layers for more precise boundaries, with applications in gland segmentation and aerial image labeling, respectively. The SharpMask~\cite{pinheiro_learning_2016} approach uses a multi-stage network to successively learn refinements of the segmented shapes.

These methods all try to alleviate the classification noise by incorporating spatial-awareness in the semantic segmentation pipeline. However, they share a common drawback as they introduce an explicit hand-crafted loss term to sharpen boundaries and spatially regularize the segmentation, either in the form of a regularization loss penalty, a heavy network modification or a graphical model post-processing. This stems from the fact that segmentation labels are often an aggregation of binary masks that have a low spatial-expressiveness. In~\cite{hayder_boundary-aware_2017}, a distance transform was introduced to allow an instance segmentation to infer shapes outside the original bounding box of the object. Indeed, the distance transform conveys proximity meaning along the edges and even further. This allows the network to learn more precise information than only ``in'' or ``out'' as would do one-hot encoding and therefore feeds cues about the spatial structure to the network.


Inspired by this recent idea, we introduce a distance transform regression loss in a multi-task learning framework, which acts as a natural regularizer for semantic segmentation. This idea was tested independently from us in~\cite{bischke_multi-task_2017}, although only for building footprint extraction using a quantized distance transform that was roughly equivalent to standard multi-class classification task. Our method is simpler as it directly works on the distance transform using a true regression. While previous methods brought additional complexity, either in the form of a hand-crafted loss function or an alternative network design, our  approach remains straightforward and fully data-driven. It requires nearly almost no network modification as it only adds a regression target, in the form of the distance transformed labels, to the original classification task.

\begin{figure*}[!t]
	\hfill
    \begin{subfigure}[t]{0.32\textwidth}
    	\includegraphics[width=\textwidth]{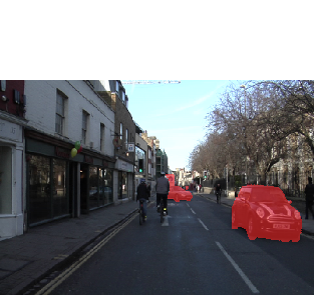}
        \caption{Car segmentation binary mask.}
    \end{subfigure}
    \hfill
    \begin{subfigure}[t]{0.32\textwidth}
    	\includegraphics[width=\textwidth]{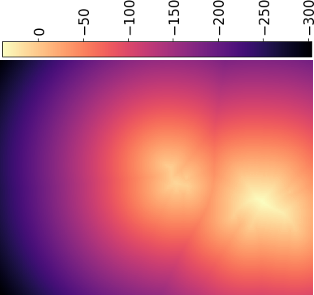}
        \caption{Signed distance transform (SDT).}
    \end{subfigure}
    \hfill
    \begin{subfigure}[t]{0.32\textwidth}
    	\includegraphics[width=\textwidth]{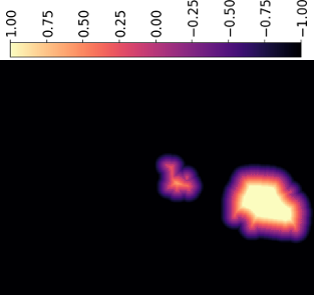}
        \caption{Truncated and normalized SDT.}
    \end{subfigure}
    \hfill
%
    \caption{Different representations of a segmentation label.}
    \label{fig:representations}
\end{figure*}

\section{Distance transform regression}

In this work, we suggest to use the signed distance transform (SDT) to improve the learning process of semantic segmentation models. The SDT transforms binary sparse masks into equivalent continuous representations. We argue that this representation is more informative for training deep networks as one pixel now owns a more precise representation of its spatial proximity with various semantic classes. We show that using a multi-task learning framework, we can train a FCN to perform both semantic segmentation by traditional classification and SDT regression, and this helps the network infer better structured semantic maps.

\subsection{Signed-distance transform}

We use the signed distance transform (SDT)~\cite{ye_signed_1988}, which assigns to each pixel of the foreground its distance to the closest background point, and to each pixel of the background the opposite of its distance to the closest foreground point. If $x_{i,j}$ are the input image pixel values and $M$ the foreground mask, then the pixels $d_{i,j}$ of the distance map are obtained with the following equation:
\begin{equation}
\forall i,j,~~~d_{i,j} =
\begin{cases}
+ min_{z \notin M}(\parallel x_{i,j} - z \parallel), & \text{if}~x_{i,j} \in M,\\
- min_{z \in M}(\parallel x_{i,j} - z \parallel), & \text{if}~x_{i,j} \notin M.\\
\end{cases}
\end{equation}

Considering that semantic segmentation annotations can be interpreted as binary masks, with one mask per class, it is possible to convert the labels into their signed-distance transform counterparts. In this work, we apply class-wise the signed Euclidean distance transform to the labels using a linear time exact algorithm~\cite{maurer_linear_2003}.

In order to avoid issues where the nearest point is outside the receptive field of the network, we clip the distance to avoid long-range spatial dependencies that would go out of the network field-of-view. The clipping value is set globally for all classes. We then normalize the SDTs of each class to constrain them in the $[-1;1]$ range. This can be seen as feeding the SDT into a non-linear saturating activation function $hardtanh$. The visual representations are illustrated in~\cref{fig:representations}. The same processing is applied to the distances estimated by the network.

\begin{figure*}[!t]
	\centering
	\includegraphics[width=0.9\textwidth]{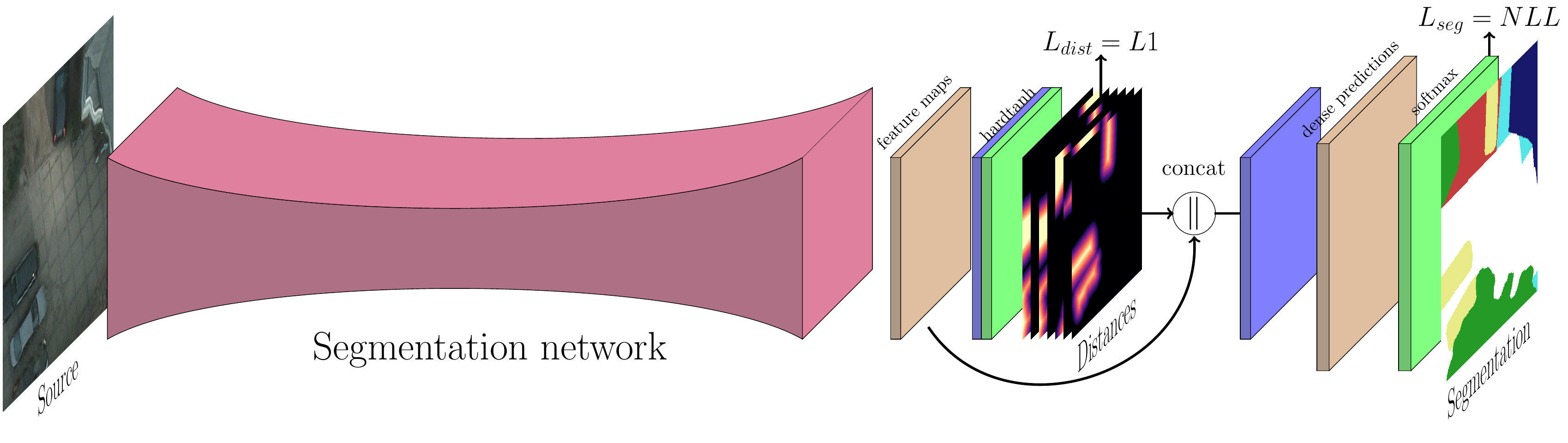}
    \caption{Multi-task learning framework by performing both distance regression and pixel-wise classification. Convolutional layers are in blue and non-linear activations are in green, while feature maps are in brown.}
    \label{fig:distance_framework}
\end{figure*}

\subsection{Multi-task learning}

Signed distance transform maps are continuous representations of the labels (classes). We can train a deep network to approximate these maps using a regression loss.

However, preliminary experiments show that training only for regression does not bring any improvement compared to traditional classification and even degrades the results. 
Therefore, we suggest to use a multi-task strategy, in which the network learns both the classification on the usual one-hot labels and the regression on all SDTs.
More precisely, we alter the network to first predict the SDTs and we then use an additional convolutional layer to fuse the last layer features and the inferred SDTs to perform the final classification. In this way, the network is trained in a cascaded multi-task fashion, where the distance transform regression is used as a proxy, i.e. an intermediate task, before classification.

Therefore, the network modification can be summarized as follows. Instead of using the last layer and feeding it into a softmax, we now use the last layer as a distance prediction. As distances are normalized between -1 and 1, these distances pass through a $hardtanh$ non-linearity. Then, we concatenate the previous layer features maps and the distance predictions to feed both into a convolutional and a softmax layer. The complete architecture is illustrated in~\cref{fig:distance_framework}.

In this work, we keep the traditional cross-entropy loss for classification, in the form of the negative log-likelihood (NLL). As our regression results are constrained in $[-1;1]$, we use the L1 loss to preserve relative errors.

Assuming that $Z_{seg}, Z_{dist}, Y_{seg}, Y_{dist}$ respectively denote the output of the segmentation softmax, the regressed distance, the ground truth segmentation labels and the ground truth distances, the final loss to be minimized is:
\begin{equation}
L = NLLLoss(Z_{seg}, Y_{seg}) + \lambda L1(Z_{dist}, Y_{dist})
\end{equation}
where $\lambda$ is an hyper-parameter that controls the strength of the regularization.

\section{Experiments}
\label{sec:experiments}

\subsection{Baselines}

We first obtain baseline results on various datasets using SegNet or PSPNet for semantic segmentation, either using the cross entropy for label classification or the L1 loss for distance regression. However, note that our method is not architecture-dependent. It consists in a straightforward modification of the end of the network that would fit any architecture designed for semantic segmentation.

SegNet~\cite{badrinarayanan_segnet_2017} is a popular architecture for semantic segmentation, originally designed for autonomous driving. It is designed around a symmetrical encoder-decoder architecture based on VGG-16~\cite{simonyan_very_2015}. The encoder results in downsampled feature maps at 1:32 resolution. These maps are then upsampled and projected in the label space by the decoder using unpooling layer. The unpooling operation replaces the decoded activations into the positions of the local maxima computed in the encoder maxpooling layers.

PSPNet~\cite{zhao_pyramid_2017} is a recent model for semantic segmentation that achieved new state-of-the-art results on several datasets~\cite{cordts_cityscapes_2016,everingham_pascal_2014}. It is based on the popular ResNet~\cite{he_deep_2016} model and uses a pyramidal module at the end to incorporate multi-scale contextual cues in the learning process. In our case, we use PSPNet, that encodes the input into feature maps at 1:32 resolution, which are then upsampled using transposed convolutions.

\subsection{Datasets}
We validate our method on several datasets in order to show its generalization capacity on multi and mono-class segmentation of both ground and aerial images.

\paragraph{ISPRS 2D Semantic Labeling}

The ISPRS 2D Semantic Labeling~\cite{rottensteiner_isprs_2012} datasets consist in two sets of aerial images.
The Vaihingen scene is comprised of 33 infrared-red-green (IRRG) tiles with a spatial resolution of 9cm/px, with an average size of $2000\times1500$px. Dense annotations are available on 16 tiles for six classes: impervious surfaces, buildings, low vegetation, trees, cars and clutter, although the latter is not included in the evaluation process.
The Potsdam scene is comprised of 38 infrared-red-green-blue (IRRGB) tiles with a spatial resolution of 5cm/px and size of $6000\times6000$px. Dense annotations for the same classes are available on 24 tiles.
Evaluation is done by splitting the datasets with a 3-fold cross-validation.

\paragraph{INRIA Aerial Image Labeling Benchmark}
The INRIA Aerial Image Labeling dataset~\cite{maggiori_can_2017} is comprised of 360 RGB tiles of $5000\times5000$px with a spatial resolution of 30cm/px on 10 cities across the globe. Half of the cities are used for training and are associated to a public ground truth of building footprints. The rest of the dataset is used only for evaluation with a hidden ground truth.

\paragraph{SUN RGB-D}
The SUN RGB-D dataset~\cite{song_sun_2015} is comprised of 10,335 RGB-D images of indoor scenes acquired from various sensors, each capturing a color image and a depth map. These images have been annotated for 37 semantic classes such as ``chairs'', ``floor'', ``wall'' or ``table'', with a few pixels unlabeled.

\paragraph{Data Fusion Contest 2015}
The Data Fusion Contest 2015~\cite{campos-taberner_processing_2016} is comprised of 7 aerial RGB images of $10,\!000\times10,\!000$px with a spatial resolution of 5cm/px on the city of Zeebruges, Belgium. A dense set of annotations on 8 classes (6 from ISPRS dataset plus ``water'' and ``boat'') is given. Two images are reserved for testing, we use one image for validation and the rest for training.

\paragraph{CamVid}
The CamVid dataset~\cite{brostow_semantic_2009} is comprised of 701 fully annotated still frames from urban driving videos, with a resolution of $360\times480$px. We use the same split as in~\cite{badrinarayanan_segnet_2017}, \textit{i.e.} 367 training images, 101 validation images and 233 test images. The ground truth covers 11 classes relevant to urban scene labeling, such as ``building'', ``road'', ``car'', ``pedestrian'' and ``sidewalk''. A few pixels are assigned to a void class that is not evaluated.


\subsection{Experimental setup}

We experiment with the SegNet and PSPNet models.

SegNet is trained for 50 epochs with a batch size of 10. Optimization is done using Stochastic Gradient Descent (SGD) with a base learning rate of 0.01, divided by 10 after 25 and 45 epochs, and a weight decay set at 0.0005. Encoder weights are initialized from VGG-16~\cite{simonyan_very_2015} trained on ImageNet~\cite{deng_imagenet_2009}, while decoder weights are randomly initialized using the policy from~\cite{he_delving_2015}.
For SUN RGB-D, in order to validate our method in a multi-modal setting, we use the FuseNet~\cite{hazirbas_fusenet_2016} architecture. This model consists in a dual-stream SegNet that learns a joint representation of both the color image and the depth map. We train it using SGD with a learning rate of 0.01 on resized $224\times224$ images.
On aerial datasets, we randomly extract $256\times256$ crops ($384\times384$ on the INRIA Labeling dataset), augmented with flipping and mirroring. Inference is done using a sliding window of the same shape with a 75\% overlap.

We train a PSP-Net on CamVid for 750 epochs using SGD with a learning rate of 0.01, divided by 10 at epoch 500, a batch size of 10 and a weight decay set at 0.0005. We extract random $224\times224$ crops from the original images and we perform random mirroring to augment the data. We fine-tune on full scale images for 200 epochs, following the practice from~\cite{jegou_one_2017}.
Our implementation of PSPNet is based on ResNet-50 pre-trained on ImageNet and does not use the auxiliary classification loss for deep supervision~\cite{zhao_pyramid_2017}.

Finally, we use median-frequency balancing to alleviate the class unbalance from SUN RGB-D and CamVid.

For a fair comparison, the same additional convolutional layer required by our regression is added to the previous classification baselines, so that both models have the same number of parameters.

All experiments are implemented using the PyTorch library~\cite{noauthor_pytorch_2016}. SDT is computed on CPU using the Scipy library~\cite{jones_scipy_2001} and cached on-memory or on-disk, which slows down training during the first epoch and uses system resources. Online SDT computation using a fast GPU implementation~\cite{zampirolli_fast_2017} would strongly alleviate those drawbacks.

\begin{table*}[!t]
\begin{tabularx}{\textwidth}{c @{} Y @{} Y @{} Y @{} Y @{} Y @{} Y @{} Y}
\toprule
Method & Dataset & OA & Roads & Buildings & Low veg. & Trees & Cars\\
\midrule
SegNet (SDT regression) & \multirow{3}{*}{Vaihingen} & 89.49 & 91.03 & 95.60 & 81.23 & 88.31 & 0.00\\
SegNet (classification) & & 90.11 $\pm$ 0.11 & 91.31 $\pm$ 0.14 & 95.59 $\pm$ 0.14 & 78.43 $\pm$ 0.22 & 89.99 $\pm$ 0.14 & \textbf{82.37} $\pm$ 1.05 \\
\textbf{SegNet (+ SDT)} & & \textbf{90.31} $\pm$ 0.12 & \textbf{91.55} $\pm$ 0.24 & \textbf{95.75} $\pm$ 0.21 & \textbf{78.80} $\pm$ 0.35 & \textbf{90.10} $\pm$ 0.11 & 81.59 $\pm$ 0.71\\
\midrule
SegNet (classification) & \multirow{2}{*}{Potsdam} & 91.85 & 94.12 & 96.09 & 88.48 & 85.44 & 96.62\\
\textbf{SegNet (+SDT)} & & \textbf{92.22} & \textbf{94.33} & \textbf{96.52} & \textbf{88.55} & \textbf{86.55} & \textbf{96.79}\\
\bottomrule
\end{tabularx}
\caption{Results on the ISPRS datasets. F1 scores per class and overall accuracy (OA) are reported.}
\label{tab:isprs_results}
\end{table*}

\begin{figure*}[!t]
\begin{subfigure}[t]{0.24\textwidth}
	\includegraphics[width=\textwidth]{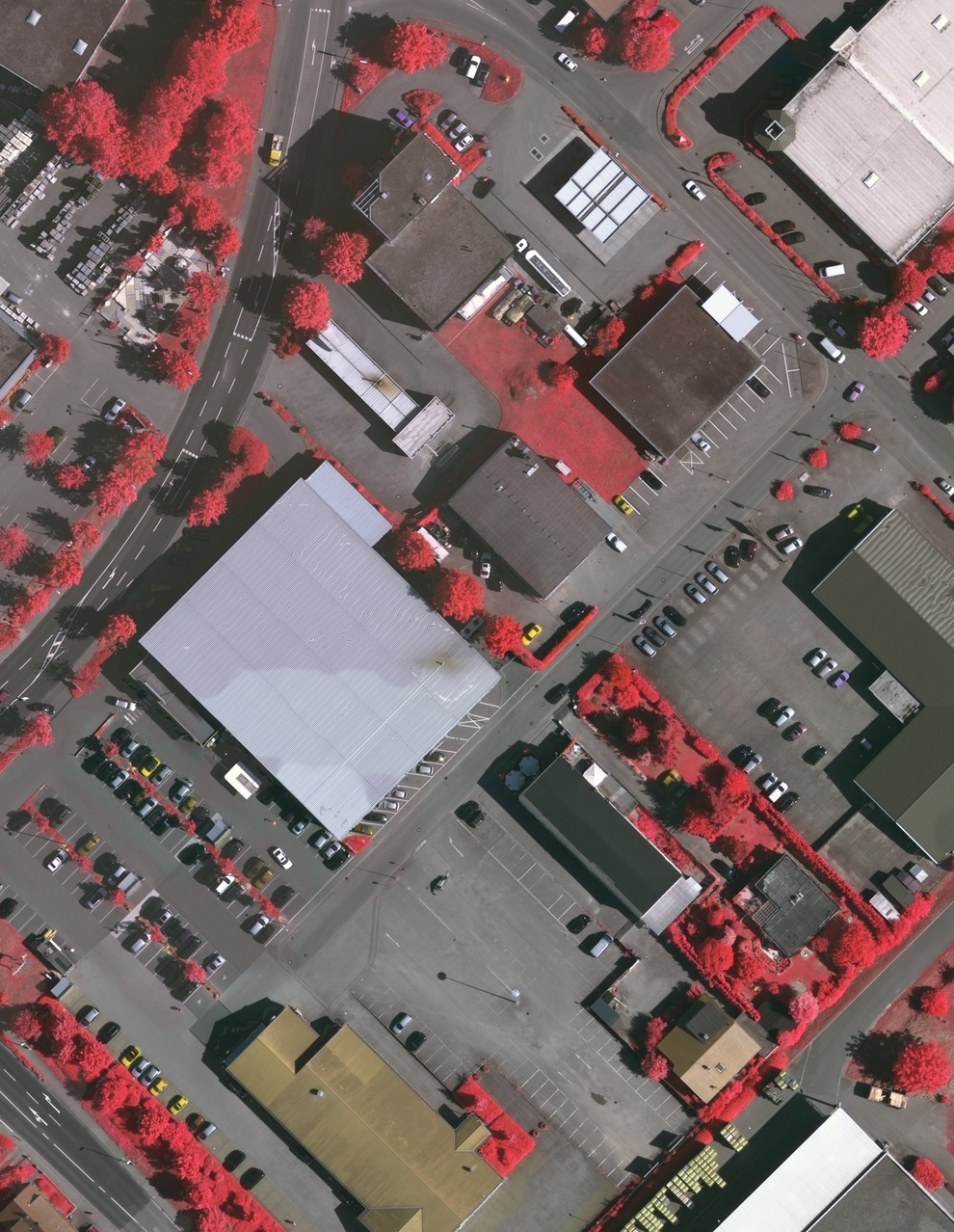}
    \caption{IRRG image}
\end{subfigure}
\begin{subfigure}[t]{0.24\textwidth}
	\includegraphics[width=\textwidth]{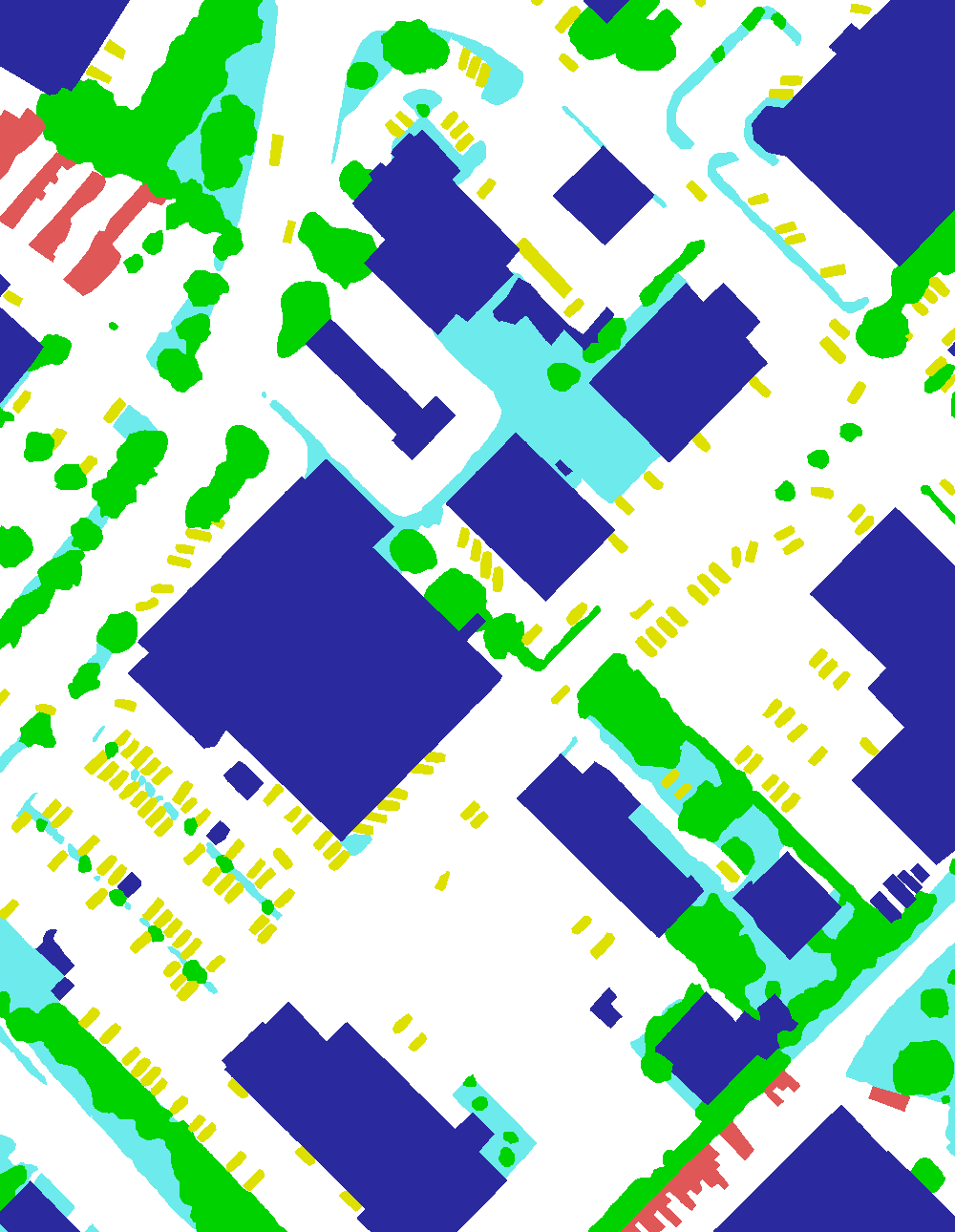}
    \caption{Ground truth}
\end{subfigure}
\begin{subfigure}[t]{0.24\textwidth}
	\includegraphics[width=\textwidth]{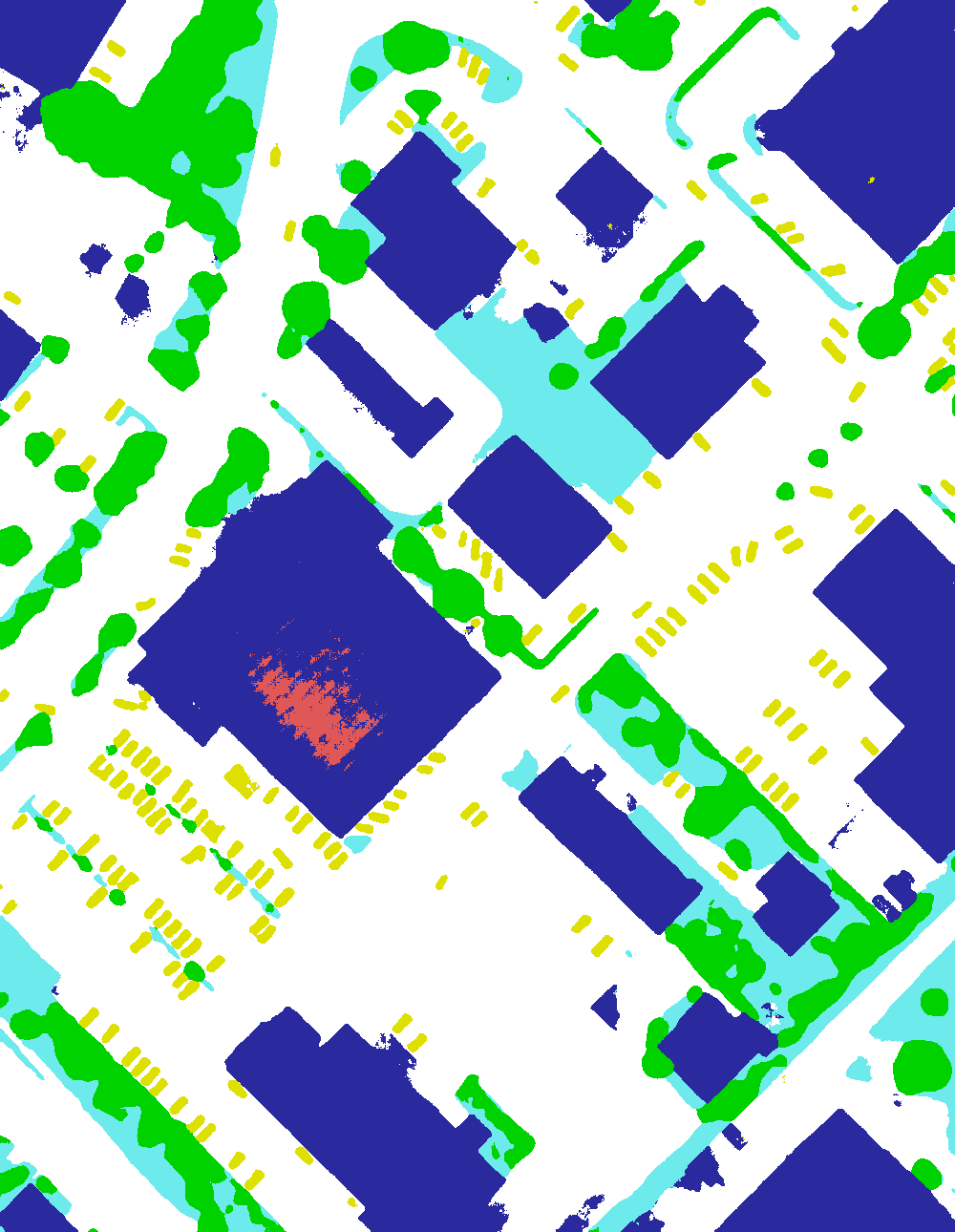}
    \caption{SegNet (classification)}
\end{subfigure}
\begin{subfigure}[t]{0.24\textwidth}
	\includegraphics[width=\textwidth]{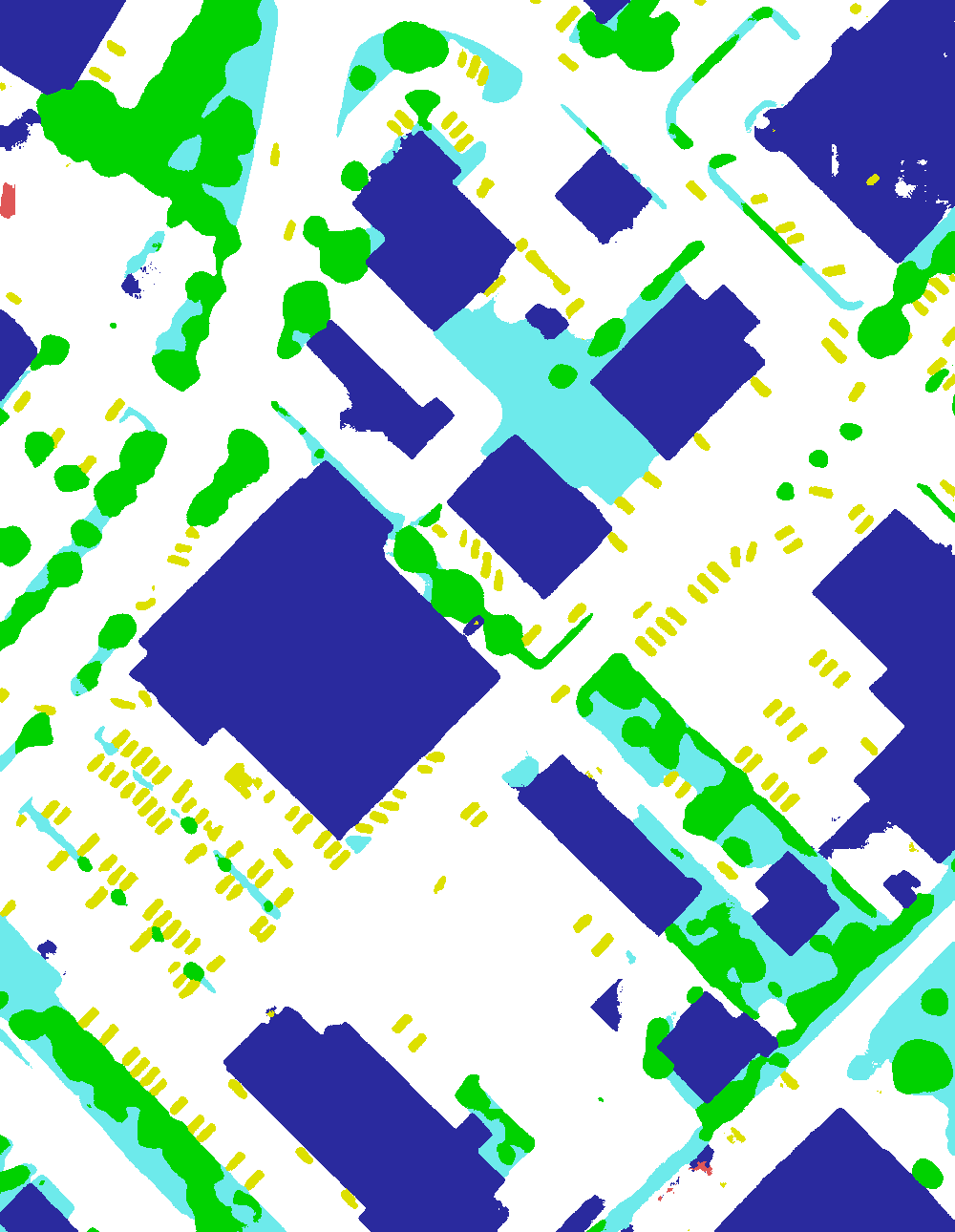}
    \caption{SegNet (multi-task)}
\end{subfigure}
\caption{Excerpt of the results on the ISPRS Vaihingen dataset. \small{Legend: white: impervious surfaces, \textcolor{Blue}{blue}: buildings, \textcolor{Cerulean}{cyan}: low vegetation, \textcolor{OliveGreen}{green}: trees, \textcolor{Dandelion}{yellow}: vehicles, \textcolor{BrickRed}{red}: clutter}, black: undefined.}
\label{fig:isprs_vaihingen}
\end{figure*}

\begin{figure}
\begin{subfigure}[t]{0.235\textwidth}
	\includegraphics[width=\textwidth]{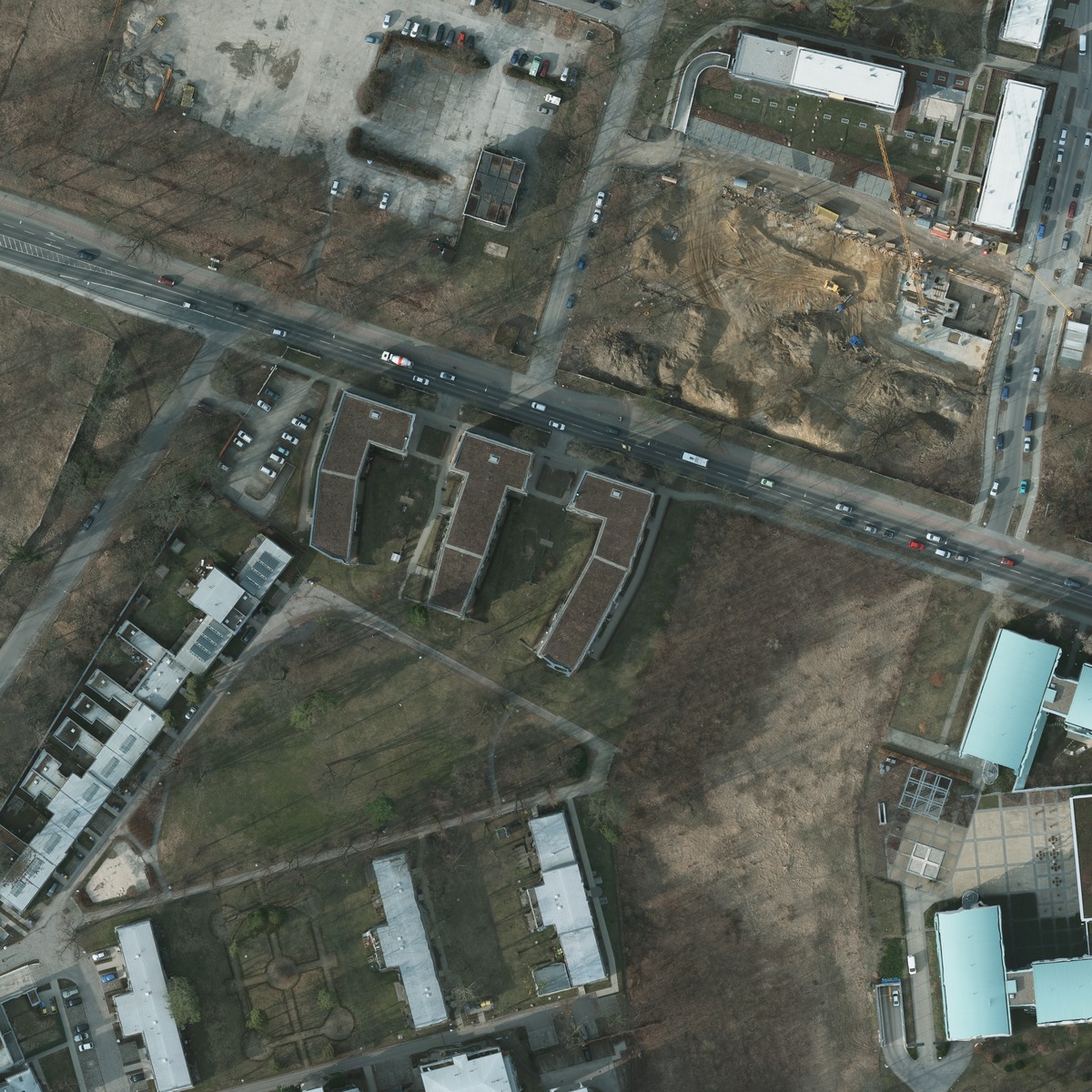}
    \caption{RGB image}
\end{subfigure}%
\hfill%
\begin{subfigure}[t]{0.235\textwidth}
	\includegraphics[width=\textwidth]{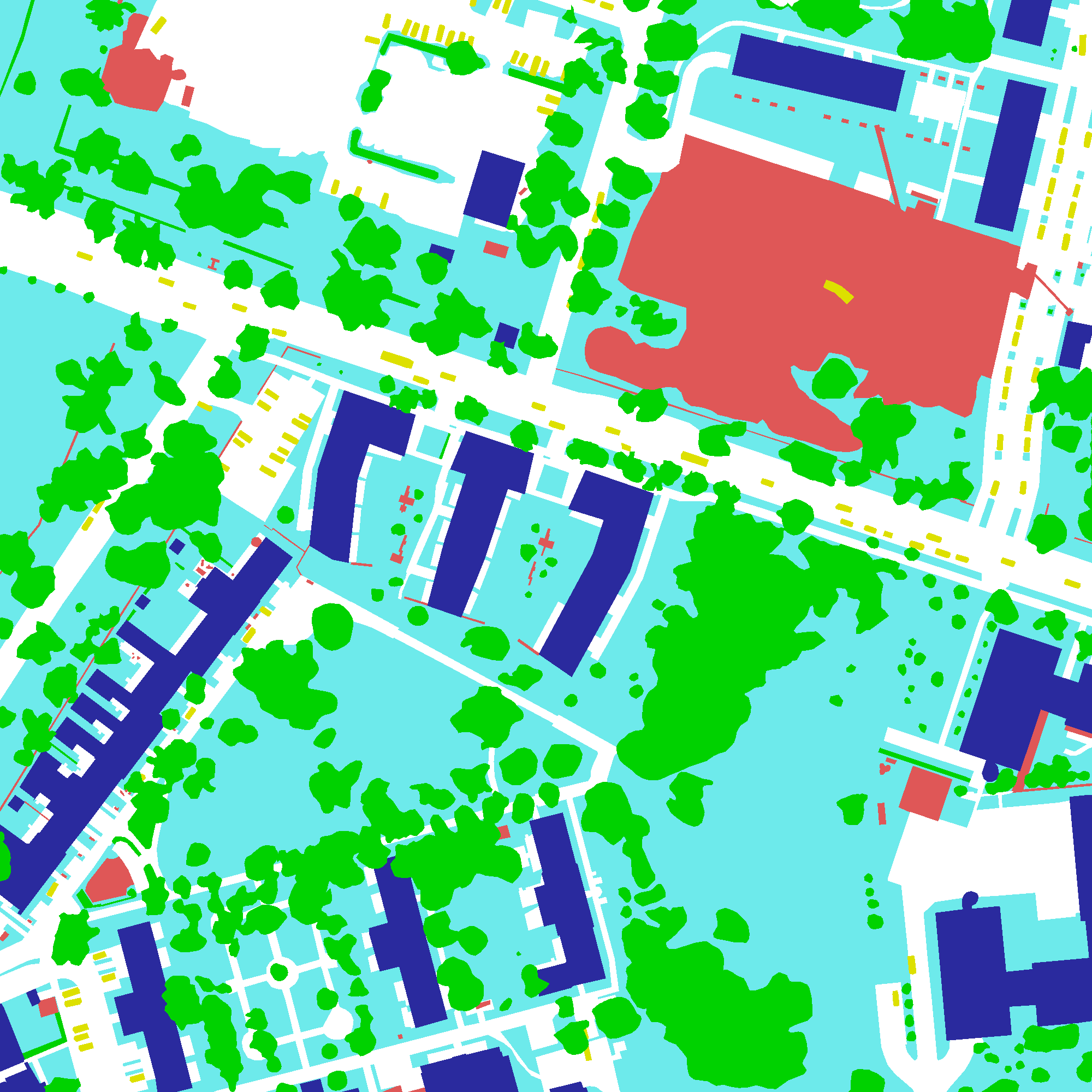}
    \caption{Ground truth}
\end{subfigure}
\begin{subfigure}[t]{0.235\textwidth}
	\includegraphics[width=\textwidth]{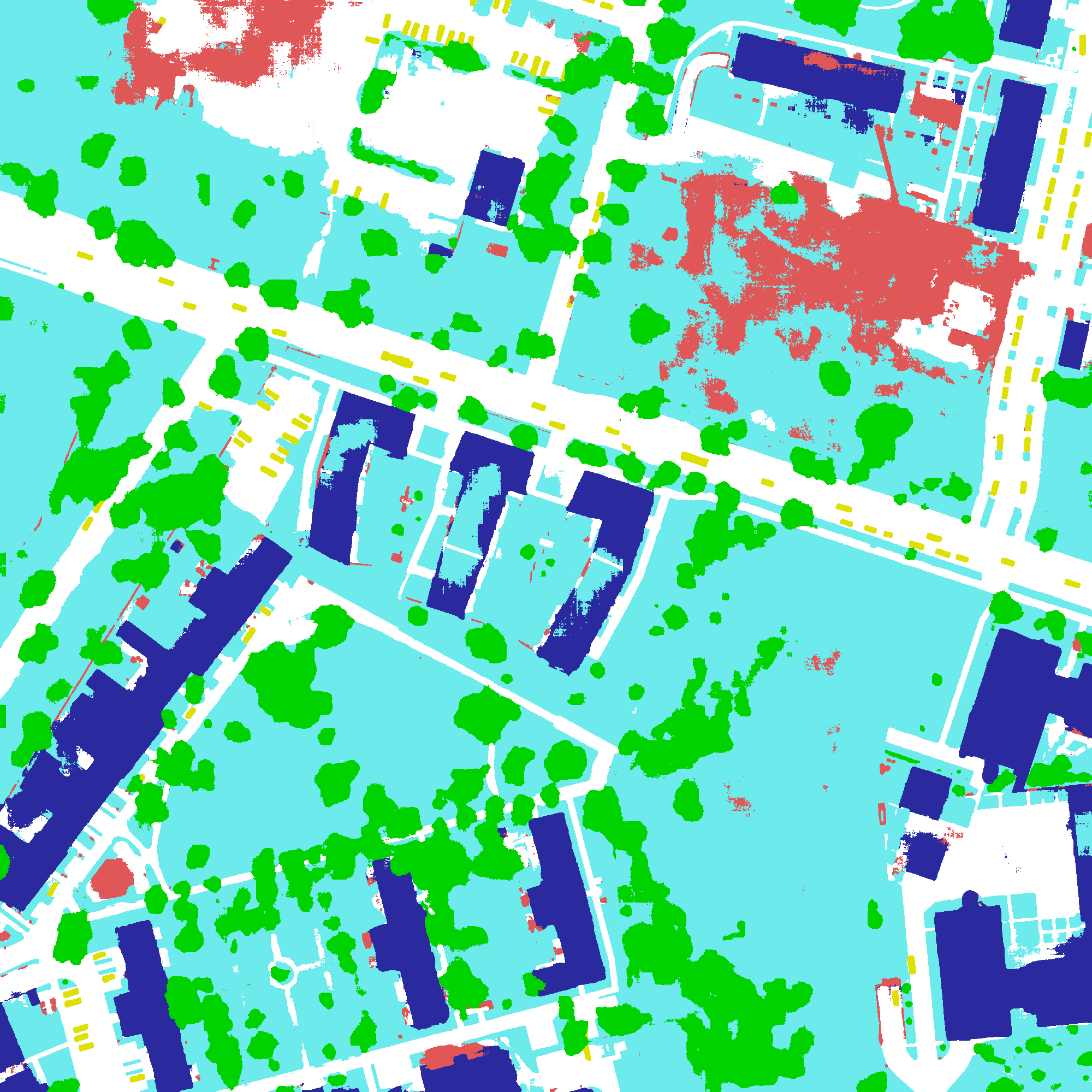}
    \caption{SegNet (classification)}
\end{subfigure}%
\hfill%
\begin{subfigure}[t]{0.235\textwidth}
	\includegraphics[width=\textwidth]{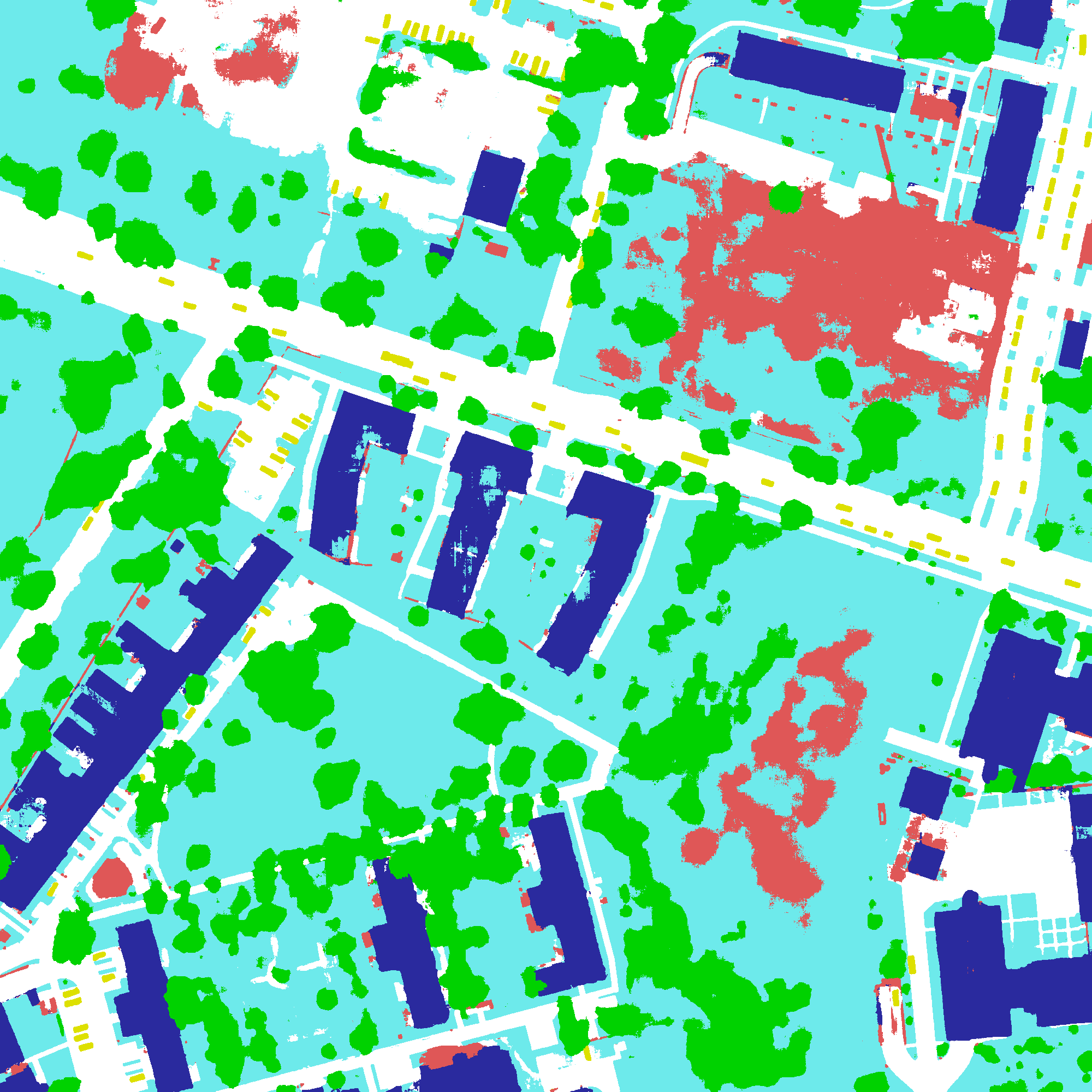}
    \caption{SegNet (multi-task)}
\end{subfigure}
\caption{Excerpt of the results on the ISPRS Potsdam dataset. \small{Legend: white: impervious surfaces, \textcolor{Blue}{blue}: buildings, \textcolor{Cerulean}{cyan}: low vegetation, \textcolor{OliveGreen}{green}: trees, \textcolor{Dandelion}{yellow}: vehicles, \textcolor{BrickRed}{red}: clutter}, black: undefined.}
\label{fig:isprs_potsdam}
\end{figure}

\subsection{Results}

\paragraph{ISPRS dataset}
The cross-validated results on the ISPRS Vaihingen and Potsdam datasets are reported in~\cref{tab:isprs_results}. For Vaihingen dataset, the validation set comprises 4 images out of 16 and 5 images out of 24 for Potsdam. All classes seem to benefit from the distance transform regression. On Potsdam, the class ``trees'' is significantly improved as the distance transform regression forces the network to better learn its closed shape, despite the absence of leaves that make the underlying ground visible from the air. Two example tiles are shown in~\cref{fig:isprs_vaihingen} and~\cref{fig:isprs_potsdam}, where most buildings strongly benefit from the distance transform regression, with smoother shapes and less classification noise. Moreover, we also tested to perform regression only on the Vaihingen dataset, which slightly improved the results on several classes, although it missed all the cars and had a negative impact overall.
It is also worth noting that our strategy succeeds while CRF did not improve classification results on this dataset as reported in~\cite{marmanis_classification_2017}.

\paragraph{INRIA Aerial Image Labeling Benchmark}

\begin{table*}[!t]
\setlength{\tabcolsep}{2pt}
\begin{tabularx}{\textwidth}{c Y Y Y Y Y Y Y}
\toprule
Method & Bellingham & Bloomington & Innsbruck & San Francisco & East Tyrol & IoU & OA\\
\midrule
Inria1  & 52.91 & 46.08 & 58.12 & 57.84 & 59.03 & 55.82 &  	93.54\\
Inria2 & 56.11 & 50.40 & 61.03 & 61.38 & 62.51 & 59.31 & 93.93\\
TeraDeep & 58.08 & 53.38 & 59.47 & 64.34 & 62.00 & 60.95 & 94.41\\
RMIT & 57.30 & 51.78 & 60.70 & 66.71 & 59.73 & 61.73 & 94.62\\
Raisa Energy & \textit{64.46} & 56.63 & 66.99 & 67.74 & 69.21 & 65.94 & 94.36\\
DukeAMLL & 66.90 & 58.48 & \textit{69.92} & \textbf{75.54} & \textit{72.34} & \textit{70.91} & \textbf{95.70}\\
NUS & 65.36 & 58.50 & 68.45 & \textit{71.17} & 71.58 & 68.36 & 95.18\\
SegNet* (classification) & 63.42 & \textit{62.74} & 63.77 & 66.53 & 65.90 & 65.04 & 94.74\\
\textbf{SegNet* (+SDT)} & \textbf{68.92} & \textbf{68.12} & \textbf{71.87} & \textit{71.17} & \textbf{74.75} & \textbf{71.02} & \textit{95.63}\\ 
\bottomrule
\end{tabularx}
\caption{Results on the test set of the INRIA Aerial Image Labeling Benchmark when our results were submitted (11/14/17). The multi-task framework consistently improves the standard SegNet results. We report the overall accuracy (OA) and the intersection over union (IoU) for each city. Best results are in \textbf{bold}, second best are in \textit{italics}.}
\label{tab:inria_results}
\end{table*}

\begin{table}[!t]
\begin{tabularx}{\linewidth}{c @{} c  c}
\toprule
Method & IoU (val) & OA (val)\\
\midrule
SegNet~\cite{bischke_multi-task_2017} & 72.57 & 95.66\\
SegNet (multi-task)~\cite{bischke_multi-task_2017} & 73.00 & 95.73\\
SegNet* (classification) & 73.70 & 95.91\\
\textbf{SegNet* (+SDT)} & \textbf{74.17} & \textbf{96.03}\\ 
\bottomrule
\end{tabularx}
\caption{Results on the validation set of the INRIA Aerial Image Labeling Benchmark for comparison to~\cite{bischke_multi-task_2017}. We report the overall accuracy (OA) and the intersection over union (IoU).}
\label{tab:inria_results_val}
\end{table}

\begin{figure*}[!t]
\begin{subfigure}{0.24\textwidth}
	\includegraphics[width=\textwidth]{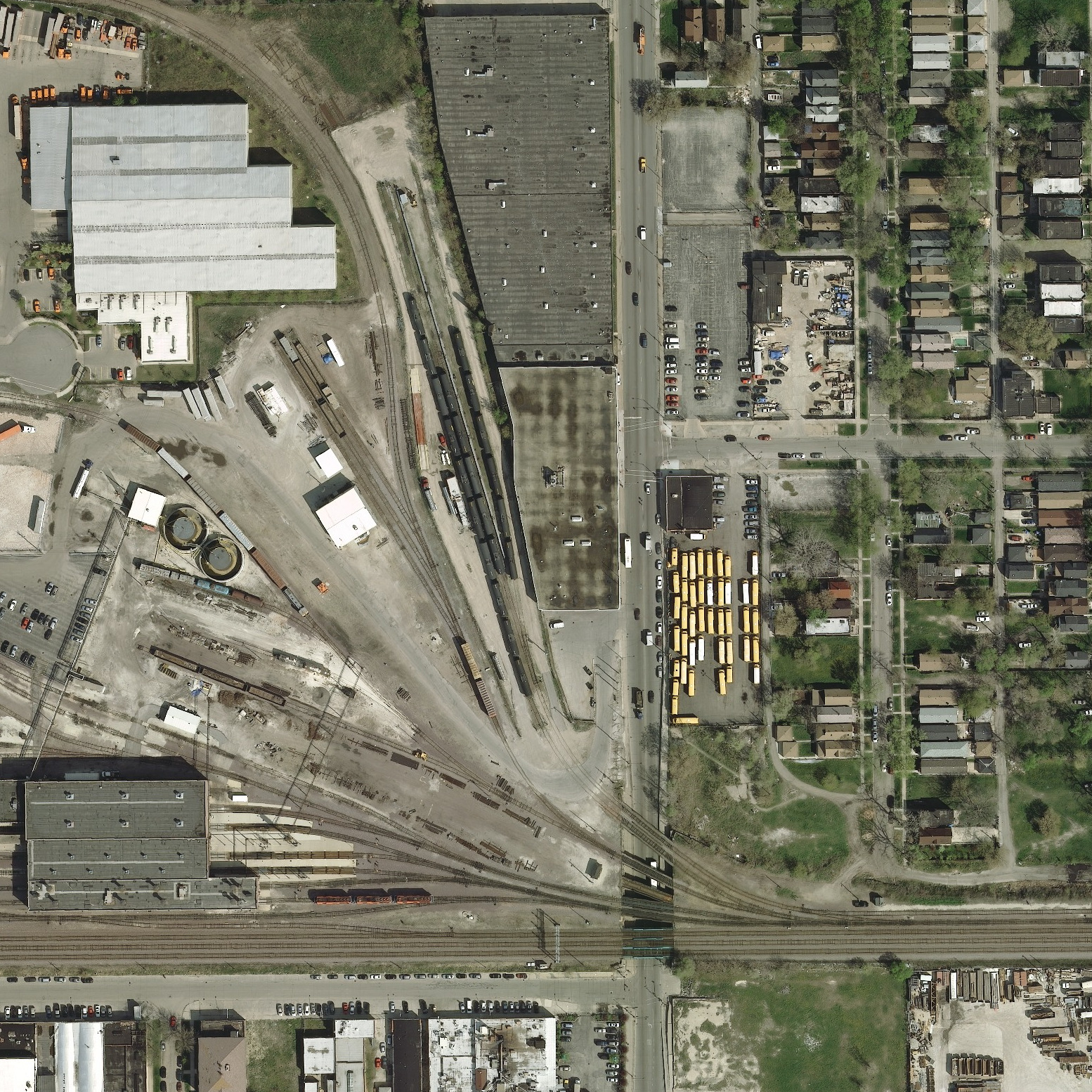}
    \caption{RGB image}
\end{subfigure}
\begin{subfigure}{0.24\textwidth}
	\includegraphics[width=\textwidth]{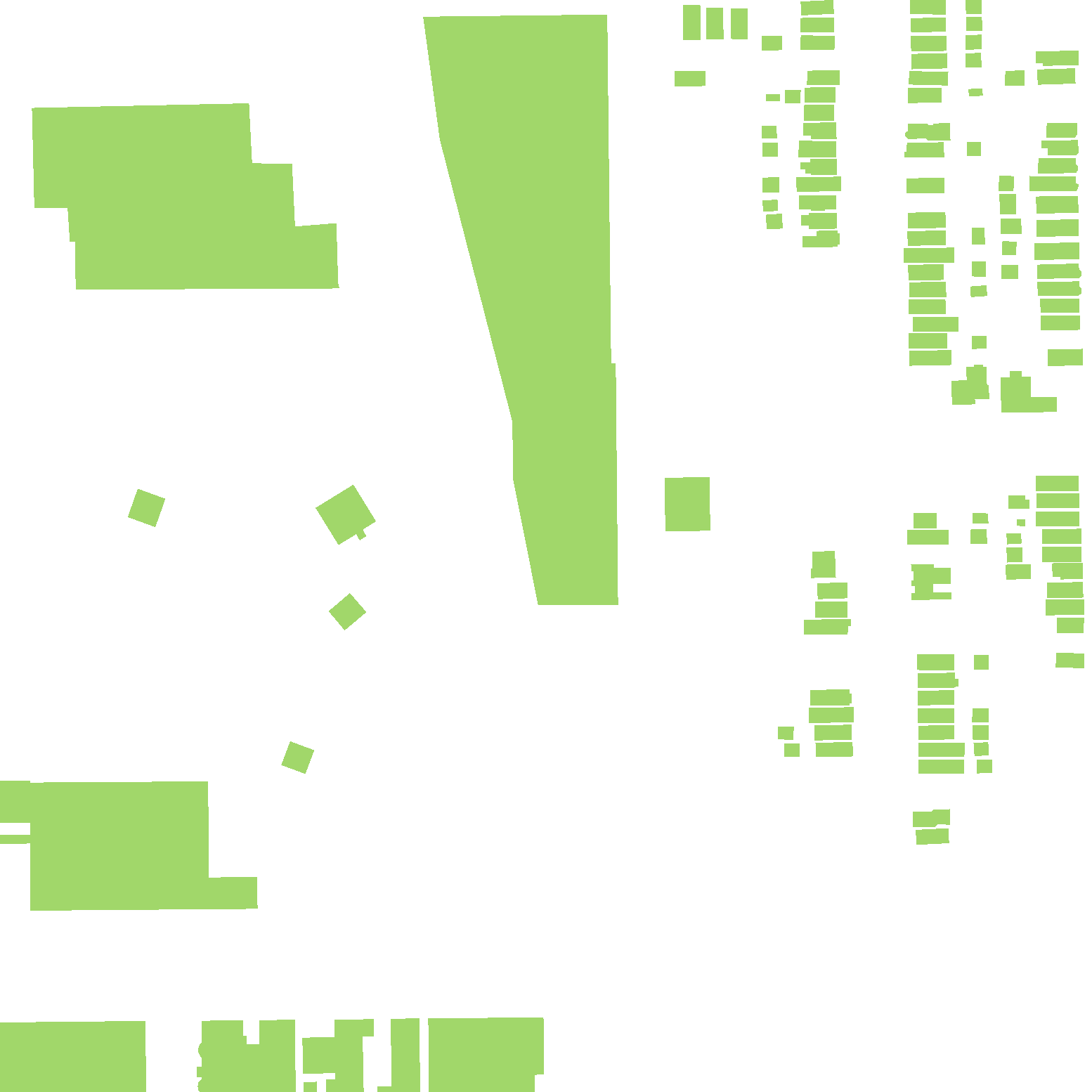}
    \caption{Ground truth}
\end{subfigure}
\begin{subfigure}{0.24\textwidth}
	\includegraphics[width=\textwidth]{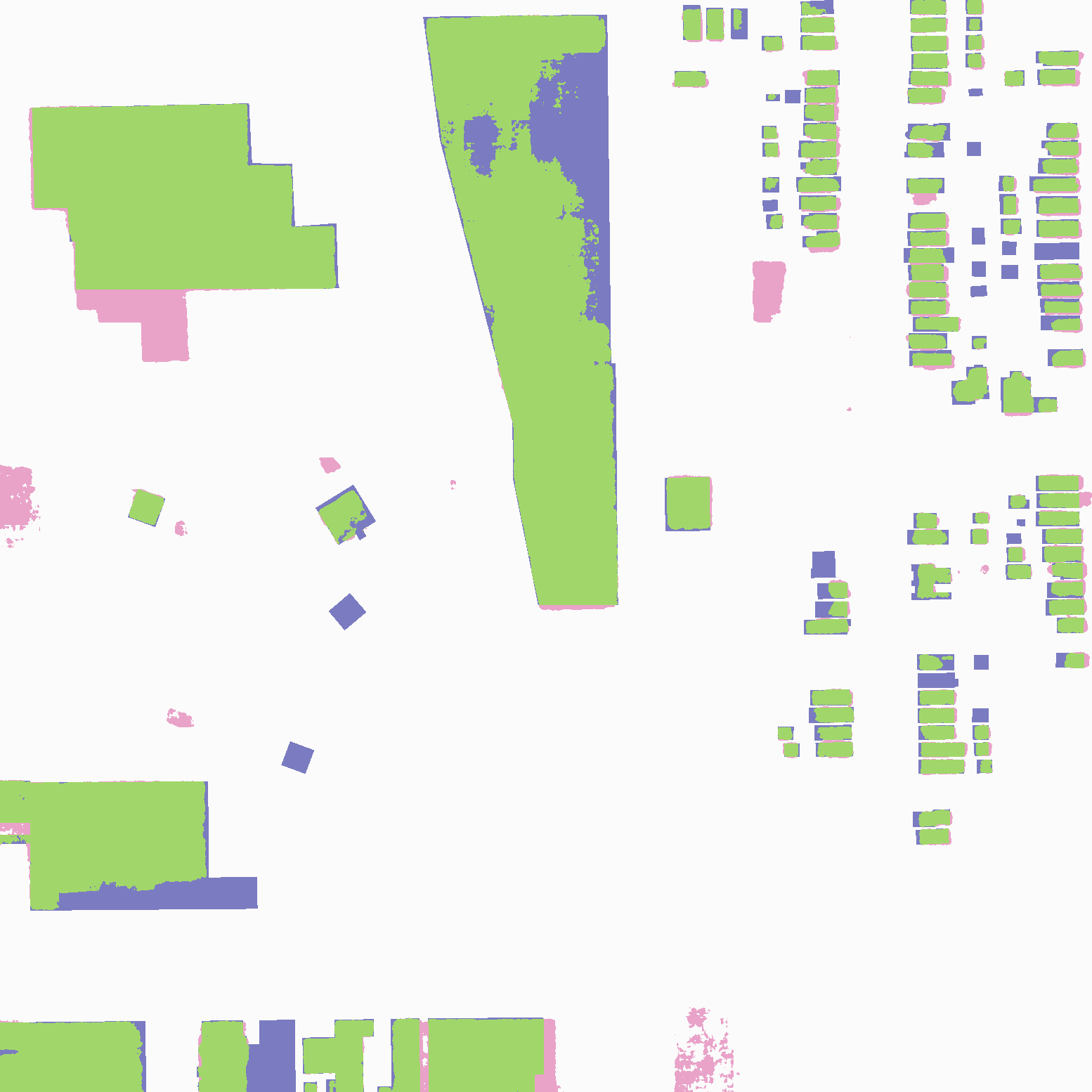}
    \caption{SegNet (standard)}
\end{subfigure}
\begin{subfigure}{0.24\textwidth}
	\includegraphics[width=\textwidth]{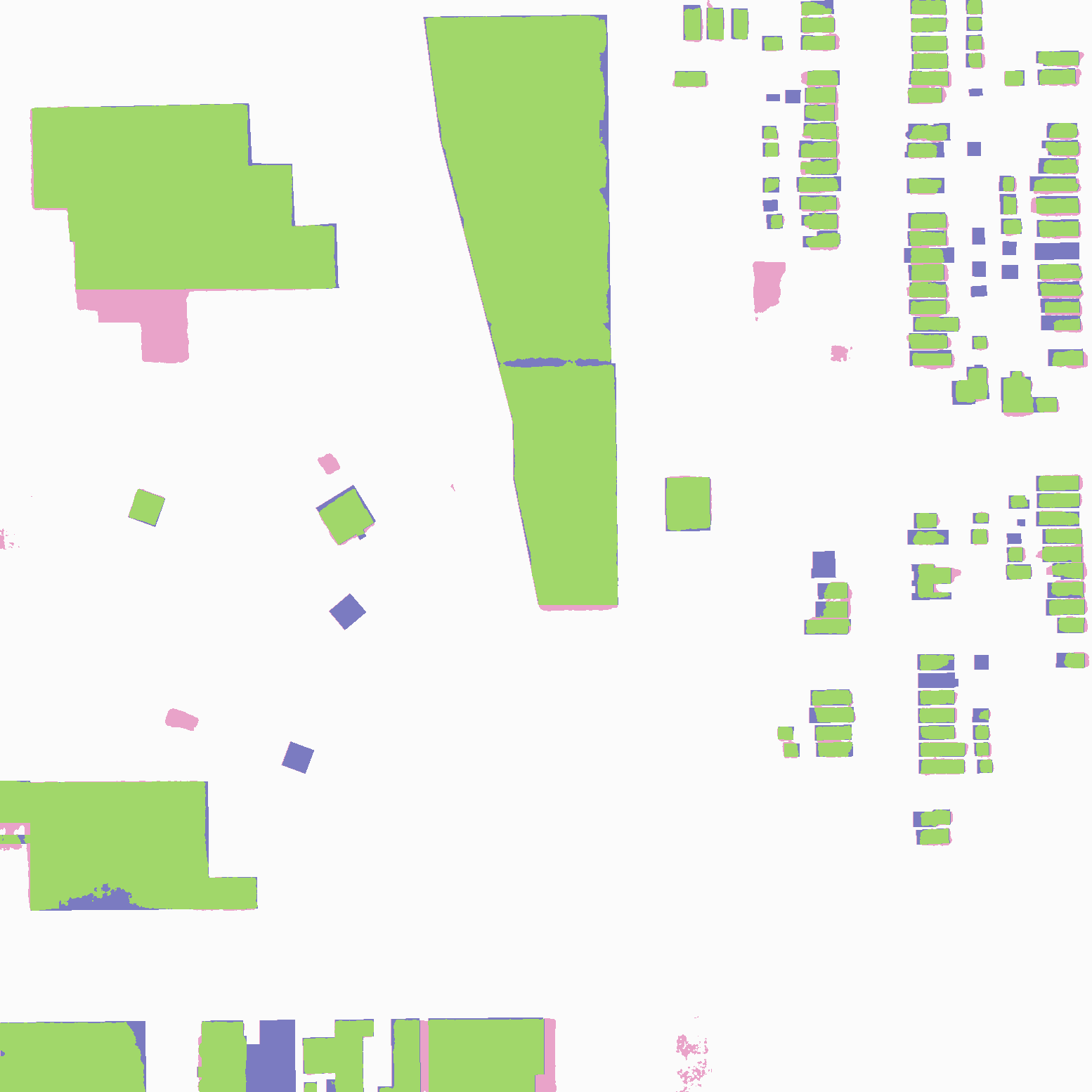}
    \caption{SegNet (multi-task)}
\end{subfigure}
\caption{Excerpt of the results on the INRIA Aerial Image Labeling dataset. Correctly classified pixels are in \textcolor{OliveGreen}{green}, false positive are in \textcolor{Lavender}{pink} and false negative are in \textcolor{RoyalBlue}{blue}. The multi-task framework allows the network to better capture the spatial structure of the buildings.}
\label{fig:inria_results}
\end{figure*}

\begin{figure*}[!t]
\begin{subfigure}{0.33\textwidth}
	\includegraphics[width=\textwidth]{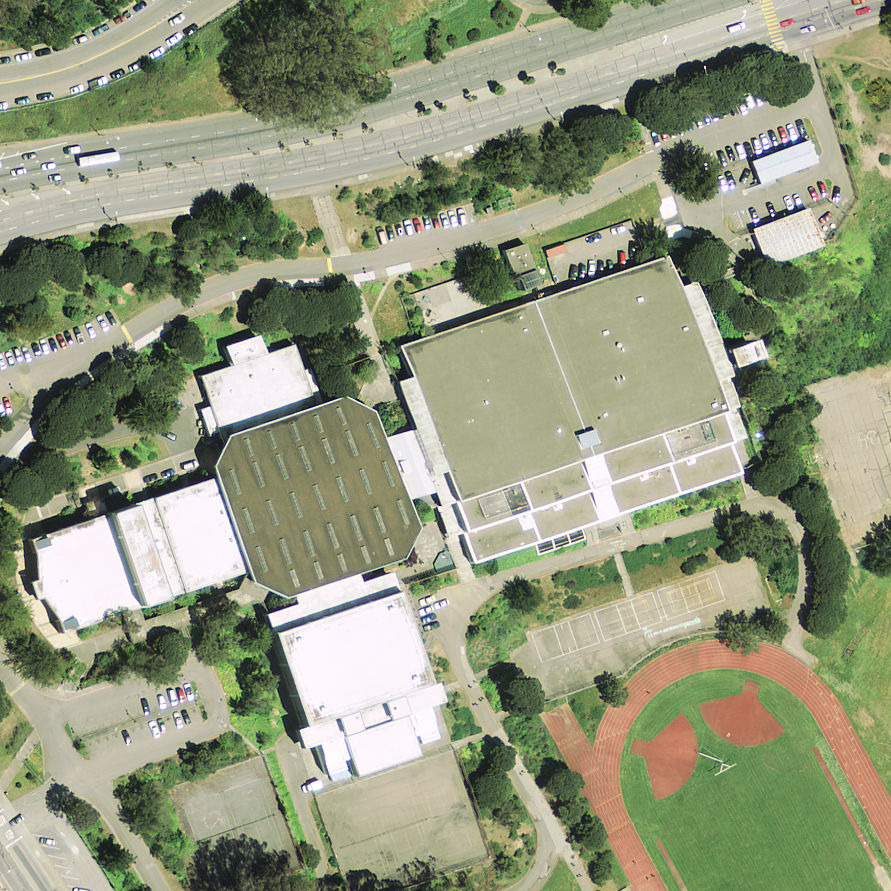}
    \caption{RGB image}
\end{subfigure}
\begin{subfigure}{0.33\textwidth}
	\includegraphics[width=\textwidth]{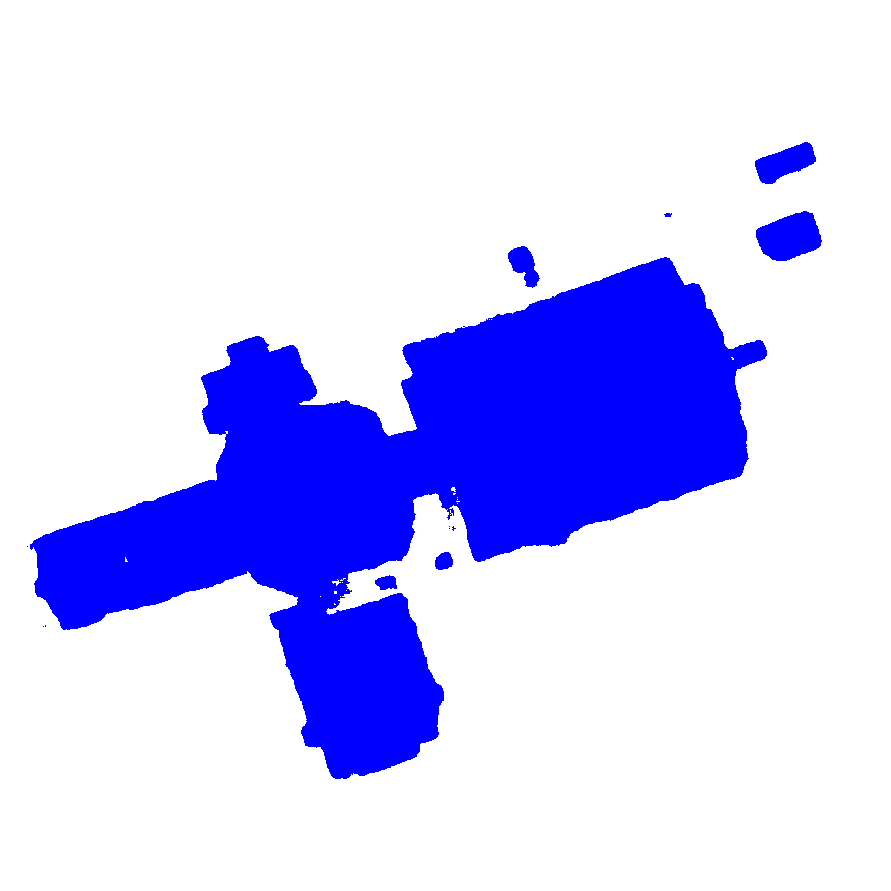}
    \caption{SegNet (standard)}
\end{subfigure}
\begin{subfigure}{0.33\textwidth}
	\includegraphics[width=\textwidth]{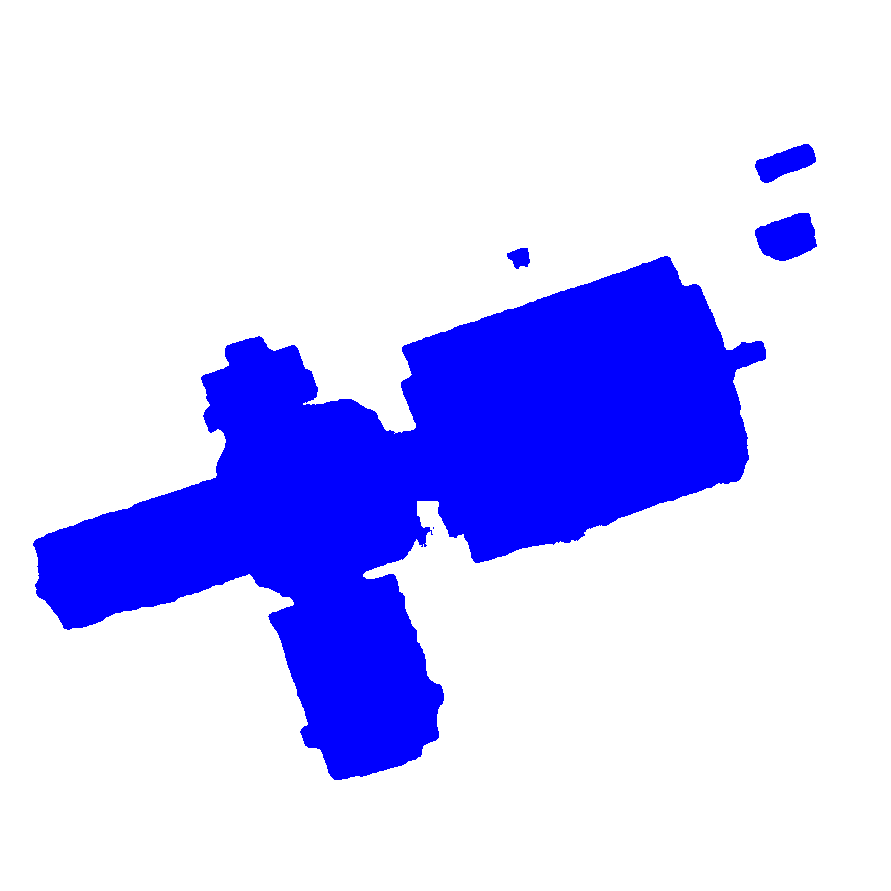}
    \caption{SegNet (multi-task)}
\end{subfigure}

\begin{subfigure}{0.33\textwidth}
	\includegraphics[width=\textwidth]{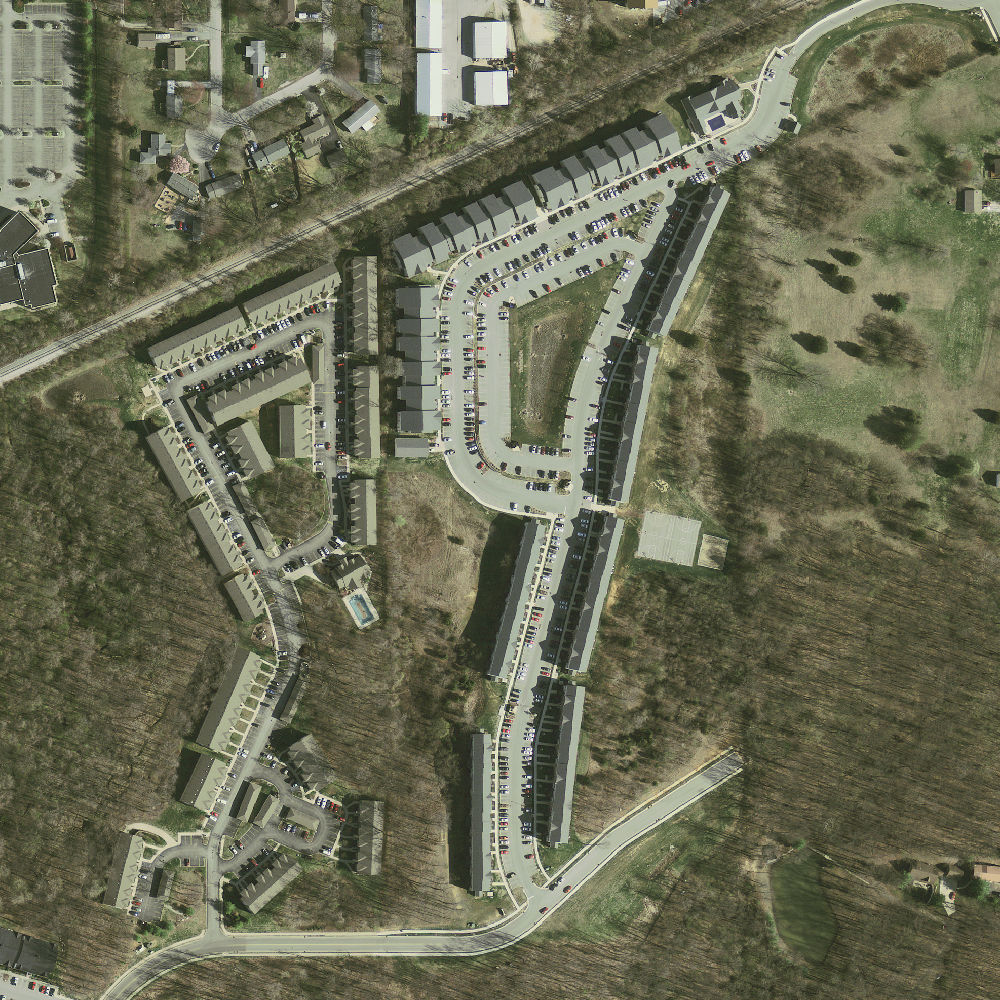}
    \caption{RGB image}
\end{subfigure}
\begin{subfigure}{0.33\textwidth}
	\includegraphics[width=\textwidth]{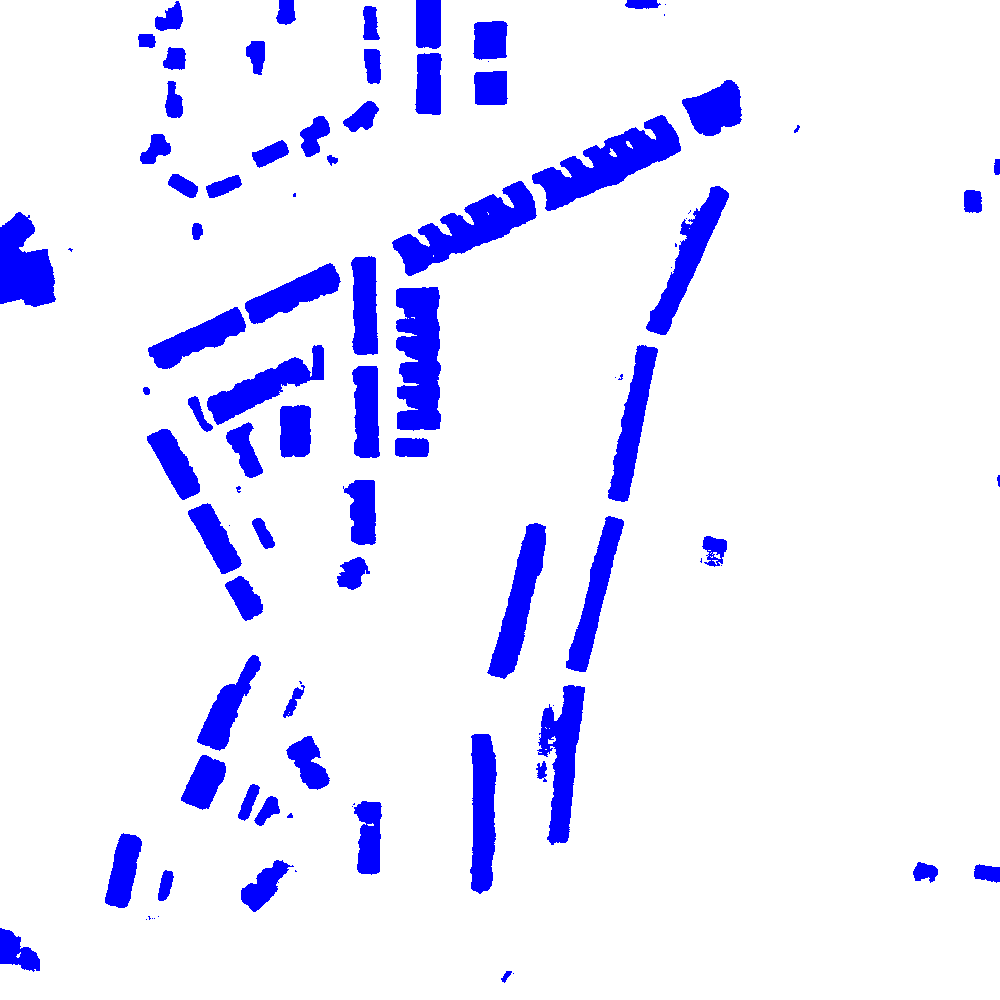}
    \caption{SegNet (standard)}
\end{subfigure}
\begin{subfigure}{0.33\textwidth}
	\includegraphics[width=\textwidth]{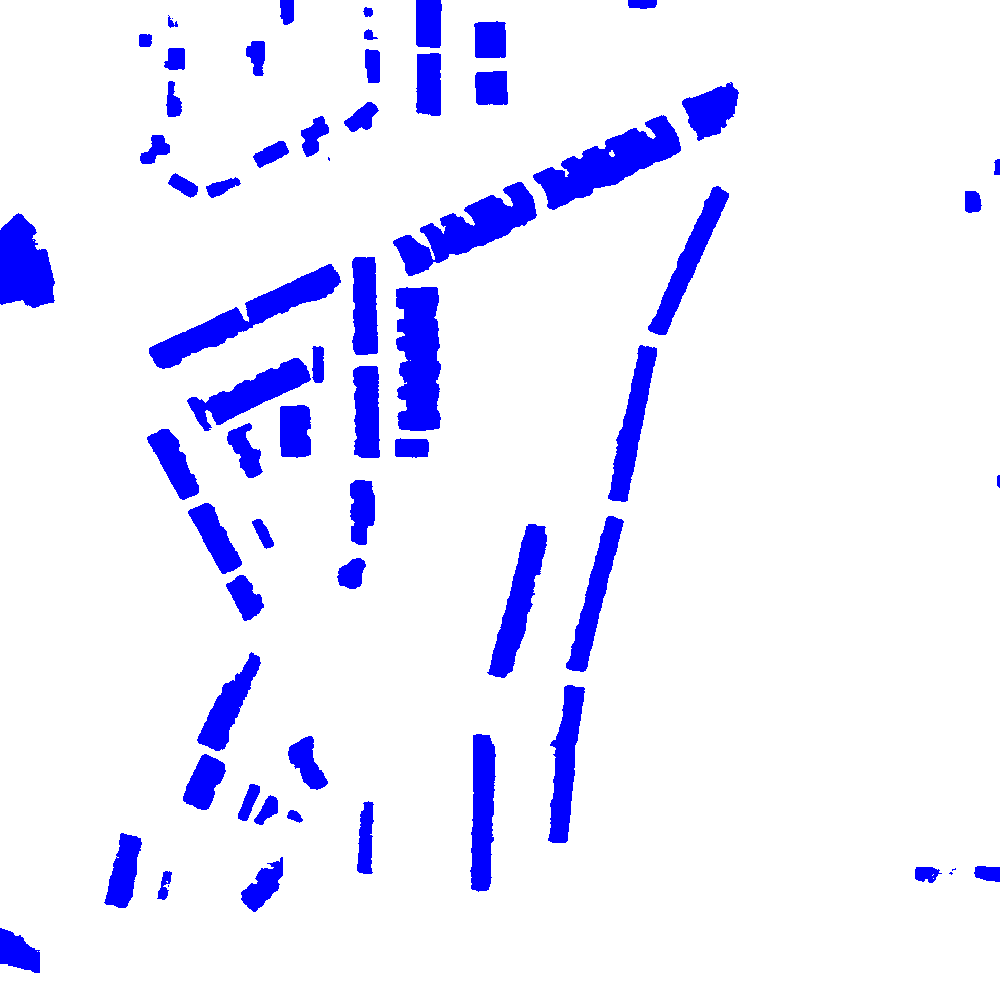}
    \caption{SegNet (multi-task)}
\end{subfigure}

\begin{subfigure}{0.33\textwidth}
	\includegraphics[width=\textwidth]{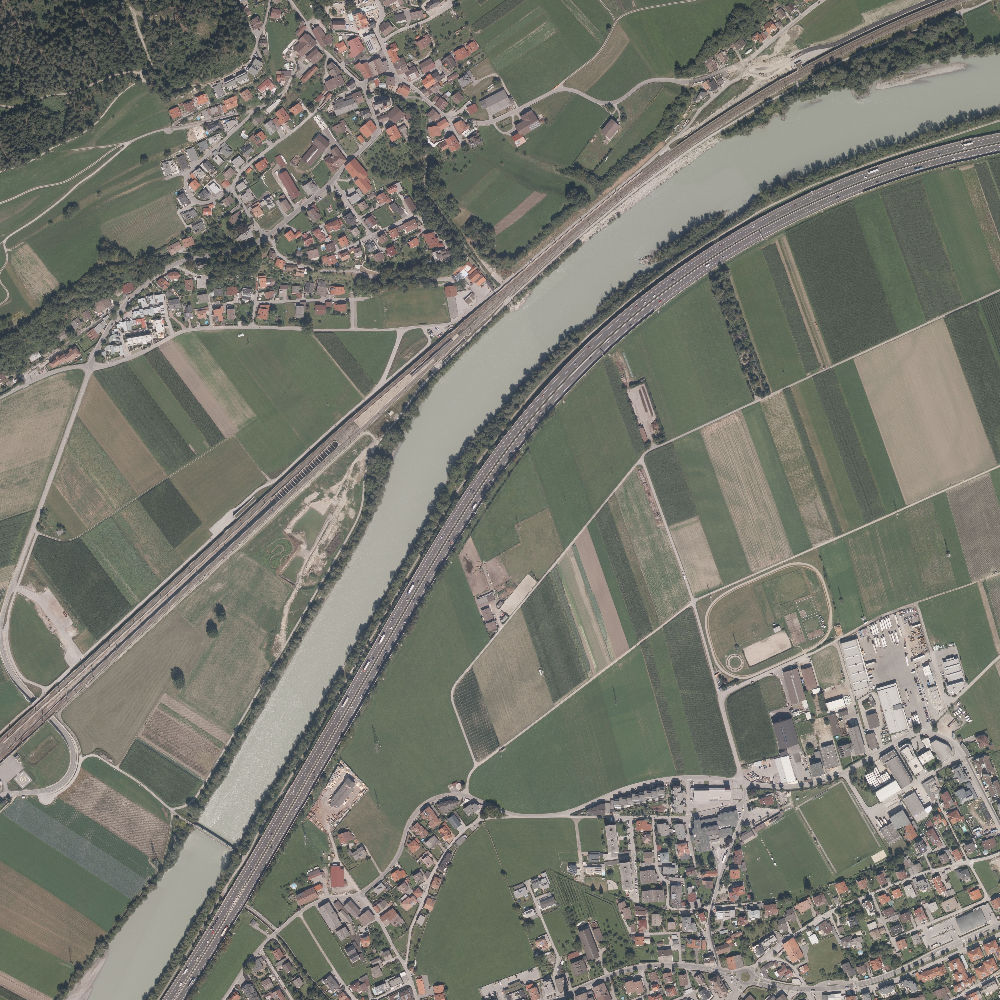}
    \caption{RGB image}
\end{subfigure}
\begin{subfigure}{0.33\textwidth}
	\includegraphics[width=\textwidth]{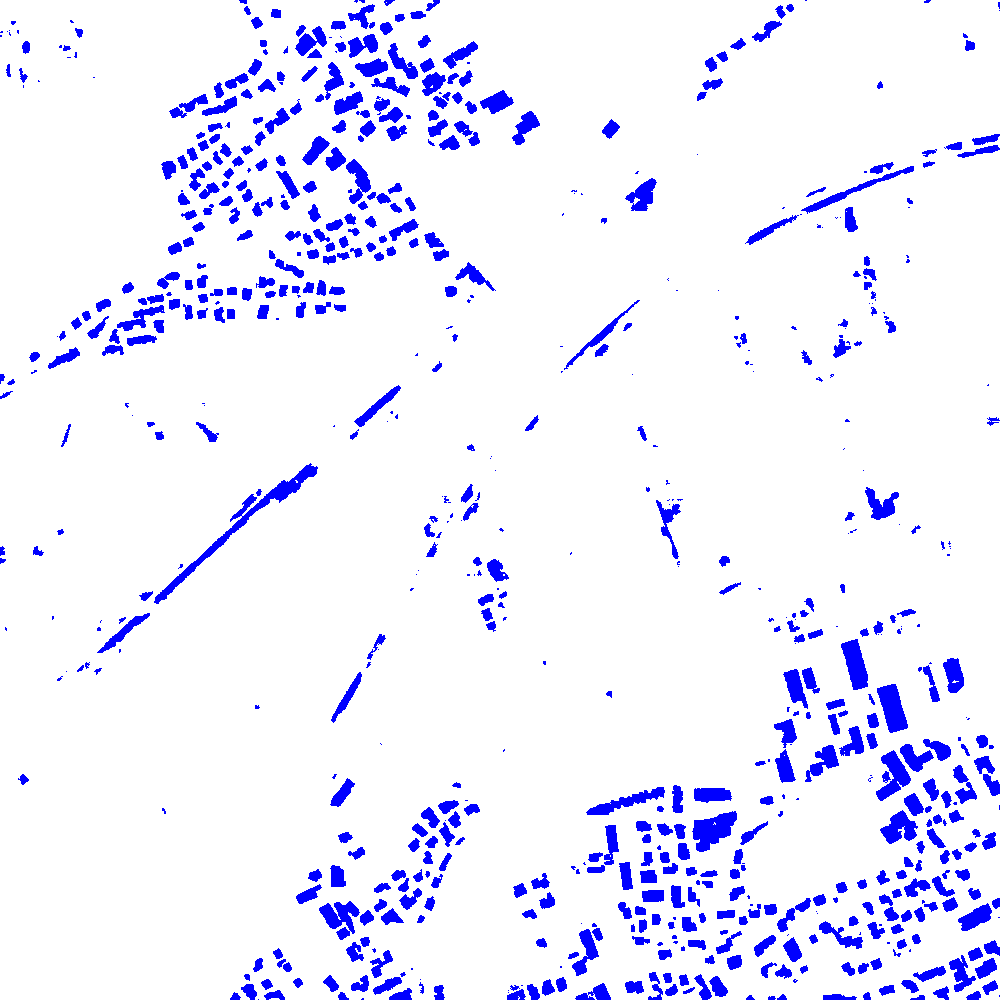}
    \caption{SegNet (standard)}
\end{subfigure}
\begin{subfigure}{0.33\textwidth}
	\includegraphics[width=\textwidth]{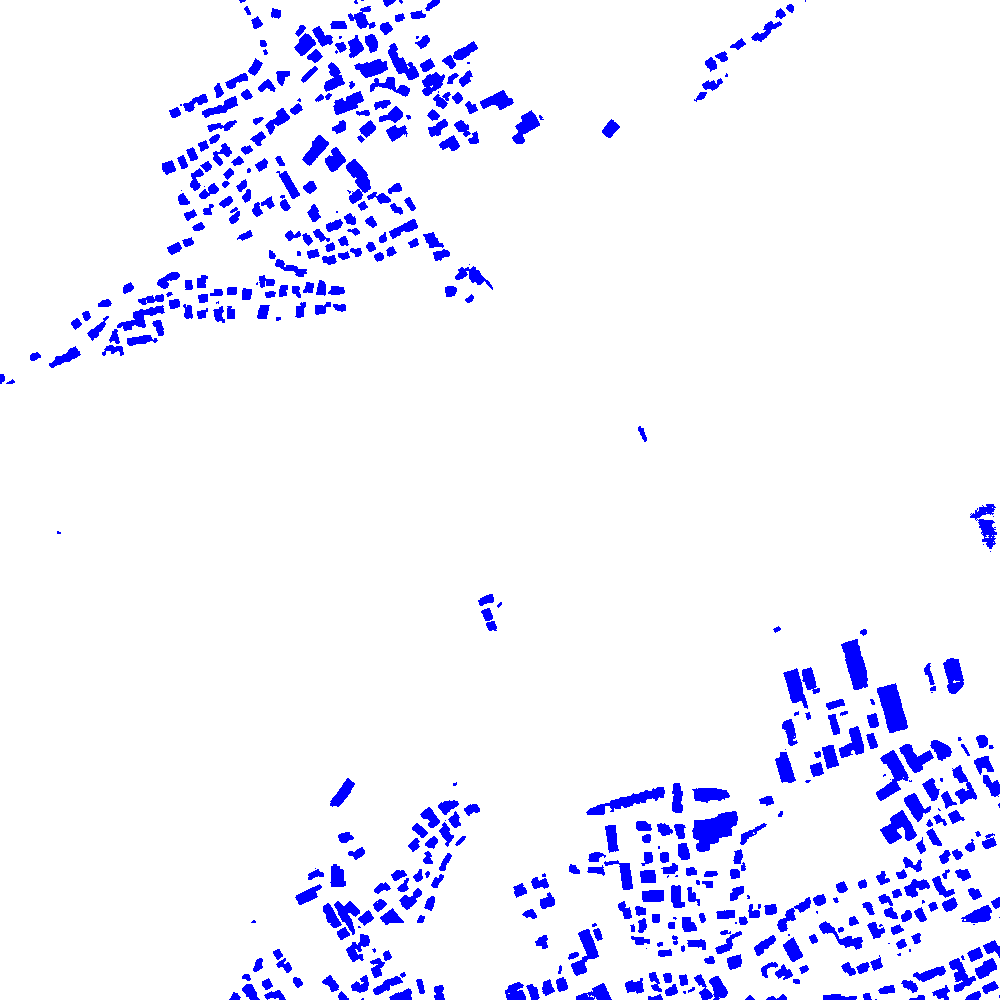}
    \caption{SegNet (multi-task)}
\end{subfigure}
\caption{Excerpt of the results on the INRIA Aerial Image Labeling test set. The multi-task framework filters out noisy predictions and cleans the predictions. Its effect is visible at multiple scales, both on a single building (more accurate shape) and on large areas (reduces the number of false positive buildings).}
\label{fig:inria_testresults}
\end{figure*}

The results on the test set of the INRIA Aerial Image Labeling benchmark are reported in~\cref{tab:inria_results}. Our results are competitive with those from other participants to the contest. Using the distance transform regression improves the intersection over union (IoU) by 0.47 and makes many errors disappear. As shown in~\cref{fig:inria_results}, multi-task prediction yields more regular building shapes and no mis-classified "holes" within the building inner part.
Although no additional buildings are detected, those that were already segmented become cleaner. Note that several missing buildings are actually false positive in the ground truth. We also present a comparison to another multi-task approach which uses a distance transform~\cite{bischke_multi-task_2017} in table~\ref{tab:inria_results_val}, this time on their custom validation set. It shows that regression on SDT is better than SDT discretization followed by classification.

\paragraph{SUN RGB-D}
\begin{table}[!t]
\begin{tabularx}{\linewidth}{c Y Y Y}
\toprule
Model & OA & AIoU & AP\\
\midrule
DFCN-DCRF~\cite{jiang_incorporating_2017} & 76.6 & 39.3 & 50.6\\
3D Graph CNN~\cite{qi_3d_2017} & - & \textit{42.0} & 55.2\\
3D Graph CNN~\cite{qi_3d_2017} (MS)  & - & \textbf{43.1} & \textit{55.7}\\
\midrule
FuseNet*~\cite{hazirbas_fusenet_2016} & \textit{76.8} & 39.0 & 55.3\\
\textbf{FuseNet* (+SDT)} & \textbf{77.0} & 38.9 & \textbf{56.5}\\
\bottomrule
\end{tabularx}
\caption{Results on the SUN RGB-D dataset on $224\times224$ images. We report the overall accuracy, average intersection over union (AIoU) and average precision (AP). We retrained our own reference FuseNet. Best results are in \textbf{bold}, second best are in \textit{italics}.}
\label{tab:sun_results}
\end{table}
We report in~\cref{tab:sun_results} test results on the SUN RGB-D dataset. Switching to the multi-task setting improves the overall accuracy and the average precision by respectively 0.33 and 1.06 points, while very slightly decreasing the average IoU. This shows that the distance transform regression also generalizes to a multi-modal setting on a dual-stream network. Note that this result is competitive with the state-of-the-art 3D Graph CNN from~\cite{qi_3d_2017} that leverages 3D cues.

\paragraph{Data Fusion Contest 2015}

\begin{table*}[!t]
\begin{tabularx}{\textwidth}{Y c c c c c c c c c}
%
\toprule
Method & OA & Roads & Buildings & Low veg. & Trees & Cars & Clutter & Boat &  Water\\
\midrule
AlexNet (patch)~{\scriptsize \cite{campos-taberner_processing_2016}} & 83.32 & 79.10 & 75.60 & 78.00 & 79.50 & 50.80 & 63.40 & 44.80 & 98.20\\
SegNet (classification) & 86.67 & \textbf{84.05} & \textbf{82.21} & 82.24 & 69.10 & 79.27 & 65.78 & \textbf{56.80} & 98.93\\
\textbf{SegNet (+SDT)} & \textbf{87.31} & 84.04 & 81.71 & \textbf{83.88} & \textbf{80.04} & \textbf{80.27} & \textbf{69.25} & 50.83 & \textbf{98.94}\\
\bottomrule
\end{tabularx}
\caption{Results on the Data Fusion Contest 2015 dataset. We report F1 scores per class and the overall accuracy (OA).}
\label{tab:dfc_results}
\end{table*}

\cref{tab:dfc_results} details the results on the Data Fusion Contest 2015 dataset compared to the best result from the original benchmark~\cite{campos-taberner_processing_2016}. Most classes benefit from the distance transform regression, with the exception of the ``boat'' class. The overall accuracy is improved by 0.64\% in the multi-task setting. Similarly to the Potsdam dataset, trees and low vegetation strongly benefit from the distance transform regression. Indeed, vegetation is often annotated as closed shapes even if it is possible to see what lies underneath. Therefore, filter responses to the pixel spectrometry can be deceptive. Learning distances forces the classifier to integrate spatial features into the decision process.

\paragraph{CamVid}

\begin{table*}[t]
\setlength{\tabcolsep}{1pt}
\begin{tabularx}{\textwidth}{Y c c c c c c c c c c c c c}
\toprule
Model & mIoU & OA & Building & Tree & Sky & Car & Sign & Road & Pedest. & Fence & Pole & Sidewalk & Cyclist\\
\midrule
SegNet~\cite{badrinarayanan_segnet_2017} & 46.4 & 62.5 & 68.7 & 52.0 & 87.0 & 58.5 & 13.4 & 86.2 & 25.3 & 17.9 & 16.0 & 60.5 & 24.8\\
DeepLab~\cite{chen_deeplab_2018} & 61.6 & -- & \textit{81.5} & \textit{74.6} & 89.0 & \textbf{82.2} & 42.3 & 92.2 & \textit{48.4} & 27.2 & 14.3 & 75.4 & 50.1\\
Tiramisu~\cite{jegou_one_2017} & 58.9 & 88.9 & 77.6 & 72.0 & \textit{92.4} & 73.2 & 31.8 & \textit{92.8} & 37.9 & 26.2 & \textit{32.6} & \textit{79.9} & 31.1\\
Tiramisu~\cite{jegou_one_2017} & \textbf{66.9} & \textbf{91.5} & \textbf{83.0} & \textbf{77.3} & \textbf{93.0} & 77.3 & \textit{43.9} & \textbf{94.5} & \textbf{59.6} & 37.1 & \textbf{37.8} & \textbf{82.2} & \textbf{50.5}\\
\midrule
PSPNet-50* (classif.) & 60.2 & 89.9 & 76.3 & 67.7 & 89.2 & 71.0 & 37.8 & 91.5 & 44.0 & 33.7 & 26.9 & 76.6 & 47.4\\
\textbf{PSPNet-50* (+ SDT)} & 60.7 & 90.1 & 76.9 & 69.7 & 88.7 & 72.7 & 38.1 & 90.6 & 44.0 & 36.6 & 27.1 & 75.6 & 47.7\\
PSPNet-50* (+ mask) & 60.0 & 89.8 & 75.6 & 67.1 & 89.6 & 71.4 & 37.3 & 92.8 & 44.4 & 36.1 & 27.6 & 75.7 & 42.6\\
\hdashline
PSPNet101* (classif.) & 60.3 & 89.3 & 74.7 & 64.1 & 89.0 & 71.8 & 36.6 & 90.8 & 44.5 & \textit{38.5} & 25.4 & 77.4 & \textit{50.3}\\
\textbf{PSPNet101* (+ SDT)} & \textit{62.2} & 90.0 & 76.2 & 66.4 &  88.8 & 78.0 & 37.6 & 90.7 & 47.2 & \textbf{40.1} & 28.6 & 78.9 & 51.2\\
\hdashline
DenseUNet* (classif.) & 59.5 & 89.6 & 75.8 & 68.6 & 90.9 & 75.3 & 37.3 & 90.0 & 42.1 & 26.5 & 30.1 & 74.1 & 43.7\\
\textbf{DenseUNet* (+ SDT)} & 61.6 & \textit{90.6} & 77.5 & 69.7 & 91.1 & \textit{78.9} & \textbf{44.0} & 90.7 & 46.9 & 23.7 & 31.6 & 77.4 & 46.2\\
\bottomrule
\end{tabularx}
\caption{Results on CamVid reporting Intersection over Union (IoU) per class, the mean IoU (mIoU) and the overall accuracy (OA). Models with an ``*'' are ours. The top part of the table shows several state-of-the-art methods, while the bottom part shows how the distance transform regression consistently improved the metrics on several models. Best results are in \textbf{bold}, second best are in \textit{italics}.}
\label{tab:camvid_results}
\end{table*}

\begin{figure*}[!t]
\centering
\begin{subfigure}{0.24\textwidth}
	\includegraphics[width=\textwidth]{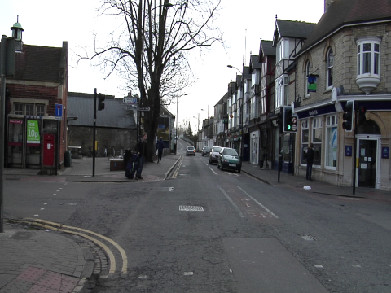}
\end{subfigure}
\begin{subfigure}{0.24\textwidth}
	\includegraphics[width=\textwidth]{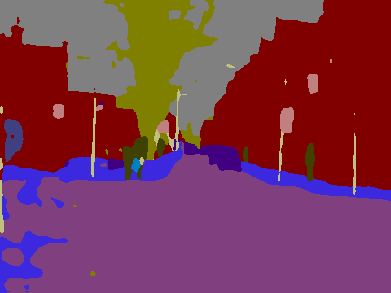}
\end{subfigure}
\begin{subfigure}{0.24\textwidth}
	\includegraphics[width=\textwidth]{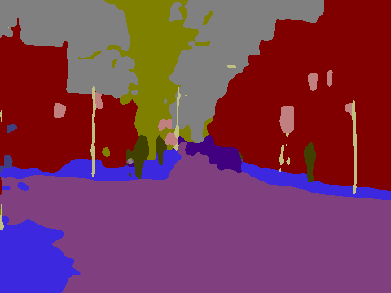}
\end{subfigure}
\begin{subfigure}{0.24\textwidth}
	\includegraphics[width=\textwidth]{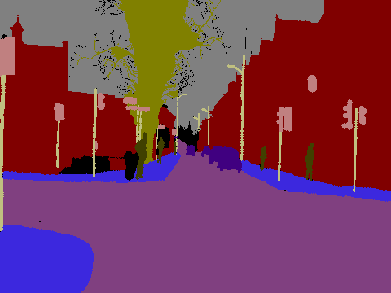}
\end{subfigure}

\begin{subfigure}{0.24\textwidth}
	\includegraphics[width=\textwidth]{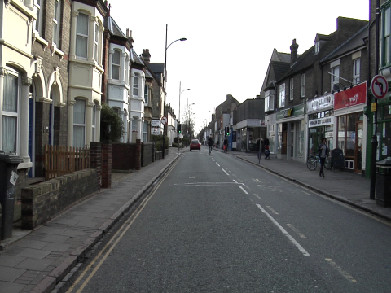}
\end{subfigure}
\begin{subfigure}{0.24\textwidth}
	\includegraphics[width=\textwidth]{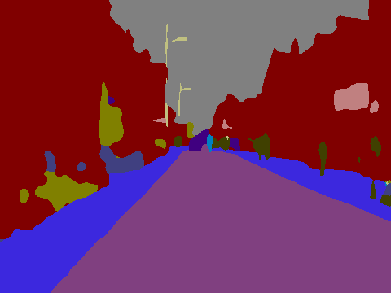}
\end{subfigure}
\begin{subfigure}{0.24\textwidth}
	\includegraphics[width=\textwidth]{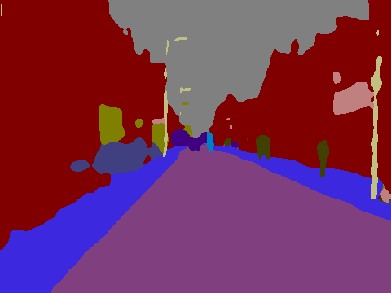}
\end{subfigure}
\begin{subfigure}{0.24\textwidth}
	\includegraphics[width=\textwidth]{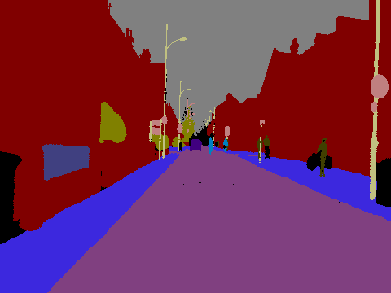}
\end{subfigure}

\begin{subfigure}{0.24\textwidth}
	\includegraphics[width=\textwidth]{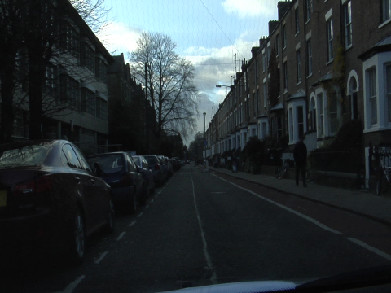}
\end{subfigure}
\begin{subfigure}{0.24\textwidth}
	\includegraphics[width=\textwidth]{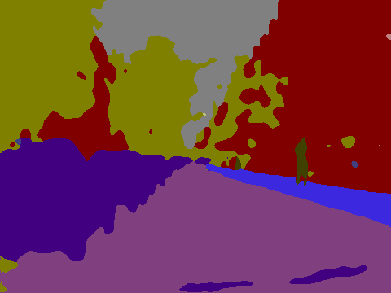}
\end{subfigure}
\begin{subfigure}{0.24\textwidth}
	\includegraphics[width=\textwidth]{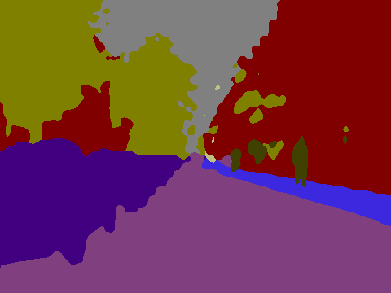}
\end{subfigure}
\begin{subfigure}{0.24\textwidth}
	\includegraphics[width=\textwidth]{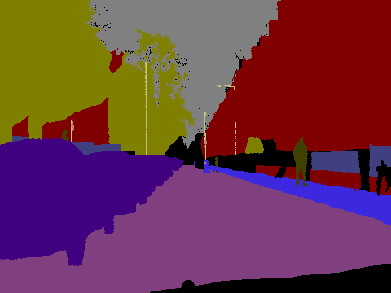}
\end{subfigure}
\caption{Example of segmentation results on CamVid using PSPNet. From left to right: RGB image, PSPNet (classification), PSPNet (multi-task), ground truth. The distance transform regression helps improve the consistency and smoothness of the sidewalks, trees, poles and traffic signs.}
\label{fig:camvid_results}
\end{figure*}

The test results on the CamVid dataset are reported in~\cref{tab:camvid_results} that also includes a comparison with other methods from the state-of-the-art, notably~\cite{jegou_one_2017}. We report here the results obtained by training two architectures: a deeper PSPNet~\cite{zhao_pyramid_2017} based on ResNet-101~\cite{he_deep_2016} and a fully convolutional DenseNet using skip connections from DenseNet~\cite{huang_densely_2017} with a UNet-inspired encoder-decoder structure~\cite{ronneberger_u-net_2015}. We use median frequency balancing for both the classification and the regression losses.

On the DenseUNet architecture, embedding the distance transform regression in the network improves the mean IoU by 2.1\% and the overall accuracy by 1.0\%, with consistent moderate improvements on all classes and significant improvements on cyclists, pedestrians and traffic signs thanks to the class balancing. On the PSPNet-101, our method improves the IoU by 1.9\% and the overall accuracy by 0.7\%, with per class metrics consistent with the other models, i.e. significant improvements on all classes except roads and sky. Some examples are shown in~\cref{fig:camvid_results} where the distance transform regression once again produces smoother segmentations.

The PSPNet baseline is competitive with those other methods and its mean IoU is improved by 0.5 by switching to the multi-task setting including the distance transform regression. Most classes benefit from the distance transform regression, with the exception of the ``road'' and ``sky'' classes. This is due to the void pixels, that are concentrated on those classes and that result in noisy distance labels.

Overall, our results are competitive with other state-of-the-art methods, with oly~\cite{jegou_one_2017} obtaining a better segmentation. However we were not able to reproduce their results and therefore unable to test the effect of the distance regression on their model, although it is expected to behave similarly to our DenseUNet reference.

\subsection{Discussion}

\paragraph{Hyperparameter tuning}

\begin{figure}[!t]
\centering
    \includegraphics[width=0.4\textwidth]{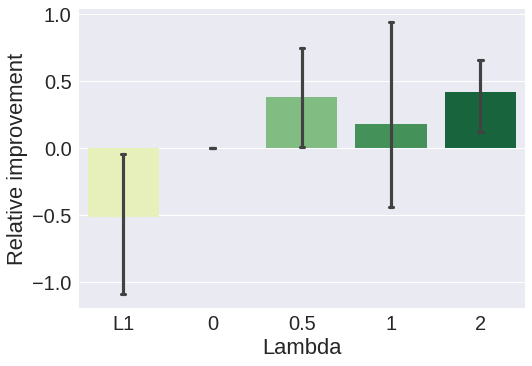}
    \caption{\textcolor{blue}{Exploration of several values for the trade-off factor of the distance transform regularization on the ISPRS Vaihingen dataset and influence on the relative improvement}. Results are obtained using a 3-fold cross-validation.}
    \label{fig:lambda_space}
\end{figure}

In order to better understand how the weight of each loss impacts the learning process, we train several models on the ISPRS Vaihingen dataset using different values for $\lambda$. This adjusts the relative influence of the distance transform regression compared to the cross-entropy classification loss. As reported in~\cref{tab:isprs_results}, we compared the regression + classification framework to both individual regression and classification. It is worth noting that SDT regression alone performs worse than the classification. This justifies the need to concatenate the inferred SDT with the last layers features in order to actually improve the performances.
Classification alone can be interpreted as $\lambda = 0$.
As can be seen in the results, incorporating the SDT regression by increasing $\lambda$ helps the network significantly. Improvements obtained with several values of $\lambda$ are detailed in~\cref{fig:lambda_space}. There are two visible peaks: one around 0.5 and one around 2. However, these two are not equivalent. The 0.5 peak is unstable and presents a high standard deviation in overall accuracy, while the $\lambda = 2$ peak is even more robust than the traditional classification.
Interestingly, this value is equivalent to rescaling the gradient from the distance regression so that its norm is approximately equal to the gradient coming from the classification. Indeed, experiments show that there is a ratio of 2 between both gradients and that they decrease roughly at the same speed during training. Therefore, it seems that better results are obtained when both tasks are given similar weights in the optimization process.
Nonetheless, all values of $\lambda$ in the test range improved the accuracy and reduced its standard deviation, making it a fairly easy hyperparameter to choose.

Finally, we also investigate the impact of using the distance transform regression compared to performing the regression on the label binary masks, which can be seen as a clipped-SDT with a threshold of 1. We experimented this on CamVid, as reported in~\cref{tab:camvid_results}, on lines PSPNet-50* (classification, +mask, +SDT). Using the L1 regression on the masks does not improve the segmentation and even worsens it on many classes. This is not surprising, as the regularization brought by the SDT regression relies on spatial cues that are absent from the binary masks.

\paragraph{Effect of the multi-task learning}

The multi-task learning incorporating the distance transform regression in the semantic segmentation model helps the network to learn spatial structures. More precisely, it constrains the network not only to learn if a pixel is in or out a class mask, but also the Euclidean distance of this pixel w.r.t the mask. This information can be critical when the filter responses are ambiguous. For example, trees from birdview might reveal the ground underneath during the winter, as there are no leaves, although annotations still consider the tree to have a shape similar to a disk. Spatial proximity helps in taking these cases into account and removing some of the salt-and-pepper classification noise that it induces, as shown on the ISPRS Vaihingen and Potsdam and DFC2015 datasets.
Moreover, as the network has to assign spatial distances to each pixel w.r.t the different classes, it also learns helpful cues regarding the spatial structure underlying the semantic maps. As illustrated in~\cref{fig:inria_results,fig:inria_testresults}, the predictions become more coherent with the original structure, with sharper boundaries and less holes when shapes are supposed to be closed.

It can be noted that multi-task is processed by concatenation of SDTs and feature maps. Indeed, concatenation with convolution generalizes the weighted sum operator. It ensures some balance in the influence of SDT and feature maps. However, other mechanisms could have been considered. For example, multiplication would intricate SDTs and feature maps. This is relevant when SDTs are well-estimated but can degrade dramatically results otherwise. For instance, in Table 1, regression fails to estimate the car SDT (which yields a null F1-score for this class) so multi-task with multiplication would also fail.

\section{Conclusion}
In this work, we looked into semantic segmentation using Fully Convolutional Networks. Semantic segmentation is the first block of many computer vision pipelines for scene understanding and therefore is a critical vision task. Despite their excellent results, FCNs often lack spatial-awareness without specific regularization techniques. This can be done using various methods ranging from graphical models to \textit{ad hoc} loss penalties.
We investigated an alternative ground truth representation for semantic segmentation tasks, in the form of the signed distance transform. We show how using both label classification and distance transform regression in a multi-task setting helps state-of-the-art networks to improve the segmentation. Especially, constraining the network to learn distance cues helps the segmentation by including spatial information. This implicit method for segmentation smoothing is fully data-driven and relies on no prior, while adding only a very low overhead to the training process. Using the distance transform regression as a regularizer, we obtained consistent quantitative and qualitative improvements in several applications of semantic segmentation for urban scene understanding, RGB-D semantic segmentation and aerial image labeling.
We argue that this method can be used straightforwardly for many use cases and could help practitioners to qualitatively improve segmentation results without using CRF or other \emph{ad hoc} graphical models.

\bibliographystyle{plainnat}
\bibliography{CVPR2018}


\end{document}